\documentclass[10pt,twocolumn,letterpaper]{article}

\usepackage{wacv}       

\usepackage[accsupp]{axessibility}
\usepackage{times}
\usepackage{epsfig}
\usepackage{graphicx}
\usepackage{amsmath}
\usepackage{amssymb}
\usepackage{booktabs}
\usepackage{graphics}
\usepackage{pgffor}
\usepackage{algorithm,algorithmic}
\usepackage[utf8]{inputenc}
\DeclareUnicodeCharacter{2212}{-}

\def\eg{\emph{e.g.}}
\def\ie{\emph{i.e.}}

\newcommand\blfootnote[1]{%
  \begingroup
  \renewcommand\thefootnote{}\footnote{#1}%
  \addtocounter{footnote}{-1}%
  \endgroup
}

\usepackage[pagebackref,breaklinks,colorlinks]{hyperref}

\usepackage[capitalize]{cleveref}
\crefname{section}{Sec.}{Secs.}
\Crefname{section}{Section}{Sections}
\Crefname{table}{Table}{Tables}
\crefname{table}{Tab.}{Tabs.}

\def\wacvPaperID{208} 

\def\confName{WACV}
\def\confYear{2024}

\begin{document}

\title{Unsupervised and semi-supervised co-salient object detection via segmentation frequency statistics}



\author{%
  \large
  Souradeep Chakraborty\textsuperscript{1,2\dag}\hskip 1em
  Shujon Naha\textsuperscript{2,3}\hskip 1em
  Muhammet Bastan\textsuperscript{2}\hskip 1em
  Amit Kumar K C\textsuperscript{2}\hskip 1em
  Dimitris Samaras\textsuperscript{1}\hskip 1em\\
  \textsuperscript{1} Stony Brook University\hskip 1em
  \textsuperscript{2} Visual Search \& AR, Amazon\hskip 1em
  \textsuperscript{3} Indiana University\hskip 1em\\
  {\tt\small \{souchakrabor,samaras\}@cs.stonybrook.edu,}
  {\tt\small \{mbastan, amitkrkc\}@amazon.com,}
  {\tt\small snaha@iu.edu}
}

\maketitle
\begin{abstract} 
In this paper, we address the detection of co-occurring salient objects (CoSOD) in an image group using frequency statistics in an unsupervised manner, which further enable us to develop a semi-supervised method. While previous works have mostly focused on fully supervised CoSOD, less attention has been allocated to detecting co-salient objects when limited segmentation annotations are  available for training. Our simple yet effective unsupervised method US-CoSOD combines the object co-occurrence frequency statistics of unsupervised single-image semantic segmentations with salient foreground detections  using self-supervised feature learning. For the first time, we show that a large unlabeled dataset e.g.  ImageNet-1k can be effectively leveraged to significantly improve unsupervised CoSOD performance. Our unsupervised model is a great pre-training initialization for our semi-supervised model SS-CoSOD, especially when very limited labeled data is available for training. To avoid propagating erroneous signals from predictions on unlabeled data, we propose a confidence estimation module to guide our semi-supervised training. Extensive experiments on three CoSOD benchmark datasets show that both of our unsupervised and semi-supervised models  outperform the corresponding state-of-the-art models by a significant margin (\eg, on the Cosal2015 dataset, our US-CoSOD model has an 8.8\% F-measure gain over a SOTA unsupervised co-segmentation model and our SS-CoSOD
model has an 11.81\% F-measure gain over a SOTA semi-supervised CoSOD model). 
\blfootnote{$^\dag$Part of this work was done during an internship at Amazon.}
\end{abstract}
\vspace{-4mm}

\section{Introduction}
\label{sec:intro}

Co-salient object detection (CoSOD) focuses on detecting co-existing salient objects in an image group, whereas salient object detection (SOD) detects the same salient objects in single images \cite{chen2020global,li2021uncertainty,liu2021samnet,piao2021mfnet,tang2021disentangled,yu2021structure,li2021salient}. CoSOD leverages the extra knowledge that the group images  share a common object by finding semantic similarities across the image regions in the group. Thus CoSOD models can localize the salient objects more accurately compared to the single image based SOD models  \cite{gao2020co,fan2021re} in such image groups. Both SOD and CoSOD are joint segmentation and detection tasks as shown in the existing literature \cite{yu2022democracy,fan2021re,fan2021group} and thus require segmentation labels. However, collecting segmentation annotations is time-consuming as well as expensive.

\begin{figure}
\centering
\includegraphics[width = 8.3cm]{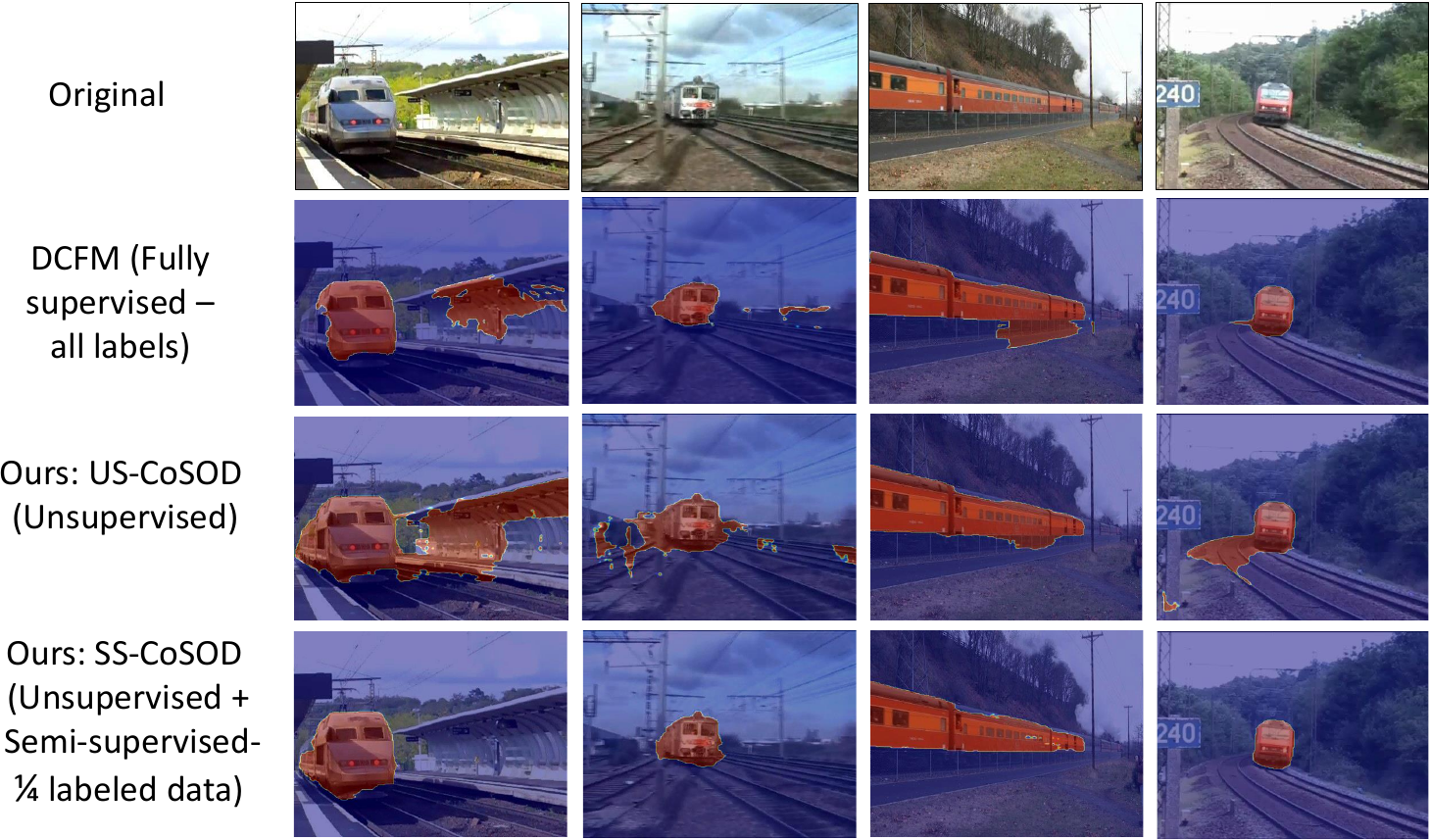}
\caption {Visualization of co-saliency detections on an image group \textit{train} from the Cosal2015 dataset \cite{zhang2016detection}. Row 1: Original image, Row 2: DCFM \cite{yu2022democracy} predictions (trained using all labeled data), Row 3:  Our unsupervised model (US-CoSOD), Row 4: Our semi-supervised model (SS-CoSOD) trained using 1/4 labeled data with unsupervised pre-training, which has comparable performance to the fully supervised DCFM trained with all labels.}
\label{fig:teaser}
\vspace{-2.5mm}
\end{figure}

This annotation requirement is a drawback for a majority of the existing CoSOD models \cite{fan2021group,zhang2021deepacg,zhang2020gradient,fan2021re,yu2022democracy} that are fully supervised. To relieve the labeling burden, some works \cite{li2018unsupervised,hsu2018co,hsu2018unsupervised} focused on unsupervised co-segmentation and co-saliency detection. Semi-supervised learning in CoSOD  \cite{semisupcosod} aims to learn an effective model from a training dataset using only a small set of labeled images along with a larger unlabeled set. Such models have an immense value in several real-world industrial applications such as in e-commerce (\eg~automatic product detection from customer review and query images of the  product without the need for manual  product annotations), content-based image retrieval, satellite imaging, bio-medical imaging, etc. However, the prediction performance of these models is significantly worse compared to the existing fully supervised models due to their inefficient use of the unlabeled data. In this paper, we first solve unsupervised CoSOD using a large unlabeled dataset and next use the unsupervised model as a pre-training initialization for our semi-supervised pipeline.

In this work, we take advantage of the recent progress in self-supervised semantic segmentation \cite{hamilton2022unsupervised} as well as self-supervised self-attention \cite{caron2021emerging} to develop a simple yet effective  unsupervised algorithm for CoSOD (US-CoSOD). As part of our unsupervised approach, we first obtain the segmentation masks of the co-occurring objects in an image group using STEGO, an off-the-shelf self-supervised semantic segmentation model ~\cite{hamilton2022unsupervised}. Next, we select the most common and salient segmentation mask (with guidance from the self-attention maps obtained from DINO \cite{caron2021emerging}, a self-supervised feature learning method) as the pseudo segmentation label for training an off-the-shelf CoSOD model in a supervised manner. We show a significant improvement in prediction performance using our methods.  However, standard training datasets are relatively small. In our paper, we introduce a more up-to-date evaluation task for unsupervised CoSOD on a set of 150K unlabeled images from the ImageNet-1k  dataset~\cite{krizhevsky2017imagenet,deng2009imagenet} (which only contains class labels without any segmentation annotation). 

Next, we show that our unsupervised model forms a strong pre-training initialization for a CoSOD model trained in a semi-supervised  manner. For this, we propose a confidence aware student-teacher architecture based semi-supervised model, SS-CoSOD. Here, we leverage the fact that in an input image group for CoSOD, we can mix the labeled and unlabeled images to effectively propagate knowledge from the labeled images to the unlabeled images in the image group via cross-region correspondences. We also introduce a confidence estimation module to block erroneous knowledge flow from inaccurate predictions on difficult unlabeled images. 
Similar to US-CoSOD, we leverage the large unlabeled ImageNet-1k  \cite{krizhevsky2017imagenet,deng2009imagenet} dataset to significantly improve semi-supervised CoSOD performance.   

In Fig.~\ref{fig:teaser}, we compare our unsupervised and semi-supervised models with DCFM \cite{yu2022democracy}, a state-of-the-art fully supervised CoSOD model. Our US-CoSOD produces segmentations comparable with DCFM and our SS-CoSOD model further improves the segmentation predictions. Our main contributions are:
\vspace{-1mm}
\begin{itemize}
  \item We propose a simple yet effective unsupervised approach for CoSOD that effectively leverages single-image semantic segmentations and self-attention maps generated using self-supervision to generate pseudo-labels for supervised training of a CoSOD model.
  
  \item For the first time, we show that CoSOD can be significantly improved using large unlabeled datasets, \eg~ImageNet-1k \cite{krizhevsky2017imagenet}. This approach helps us achieve state-of-the-art results for unsupervised CoSOD.
  
  \item We propose a novel approach for semi-supervised CoSOD by effectively propagating knowledge from a limited labeled  set to a much larger unlabeled set via  confidence estimation and cross-region correspondence between the labeled and the unlabeled sets.
\end{itemize}

\section{Related Work}
\label{sec:related_work}
\textbf{Co-salient object detection:}
Graphical models are used to model pixel relationships in an image group  \cite{hu2021multi,jiang2020co,jiang2019multiple,jiang2019unified,wei2019deep,zhang2020adaptive}, followed by mining co-salient objects with consistent features. Some works used additional object saliency information to first mine the salient objects and then implement CoSOD \cite{jin2020icnet,zhang2021summarize,zhang2020coadnet}. Other works compute the shared attributes among input images \cite{fan2021group,zhang2021deepacg,zhang2020gradient,su2023unified,le2017co,zhu2023co,fan2021re,li2023discriminative,ge2022tcnet} and supplement semantic information with classification information. The surveys \cite{cong2018review,fan2020taking,zhang2018review} provide more information on CoSOD. DCFM \cite{yu2022democracy} mines co-salient features with democracy while reducing   background interference. We use DCFM as the backbone network in our study.

\textbf{Unsupervised segmentation:} Several unsupervised semantic segmentation approaches
use self-supervised feature learning techniques  \cite{ji2019invariant,li2021contrastive,van2020scan,cho2021picie,wang2022self}.  
Recently, STEGO \cite{hamilton2022unsupervised} showed that semantically correlated dense features from unsupervised feature learning frameworks can help distill unsupervised features into high-quality semantic labels. We use this model as a component in our unsupervised pipeline. In \cite{yin2022transfgu}, the semantic categories obtained using self-supervised learning  are  mapped to pixel-level  features via class activation maps, which serve as pseudo labels for training.  
Some papers solve unsupervised co-segmentation \cite{jerripothula2016image,li2018unsupervised,hsu2018co,amir2021deep,chakraborty2015site} and CoSOD  \cite{zhang2016detection,hsu2018unsupervised}. Li et al. \cite{li2018unsupervised} proposed an unsupervised co-segmentation model by ranking image complexities using saliency maps. Hsu et  al.~\cite{hsu2018co} developed an unsupervised co-attention based model for object co-segmentation. The same authors presented an unsupervised graphical model for CoSOD  \cite{hsu2018unsupervised} that jointly solves single-image saliency and object co-occurrence. Our US-CoSOD  outperforms all of these unsupervised models.

 \begin{figure*}[t]
\centering
\includegraphics[width = 15.7cm,height = 8.7cm]{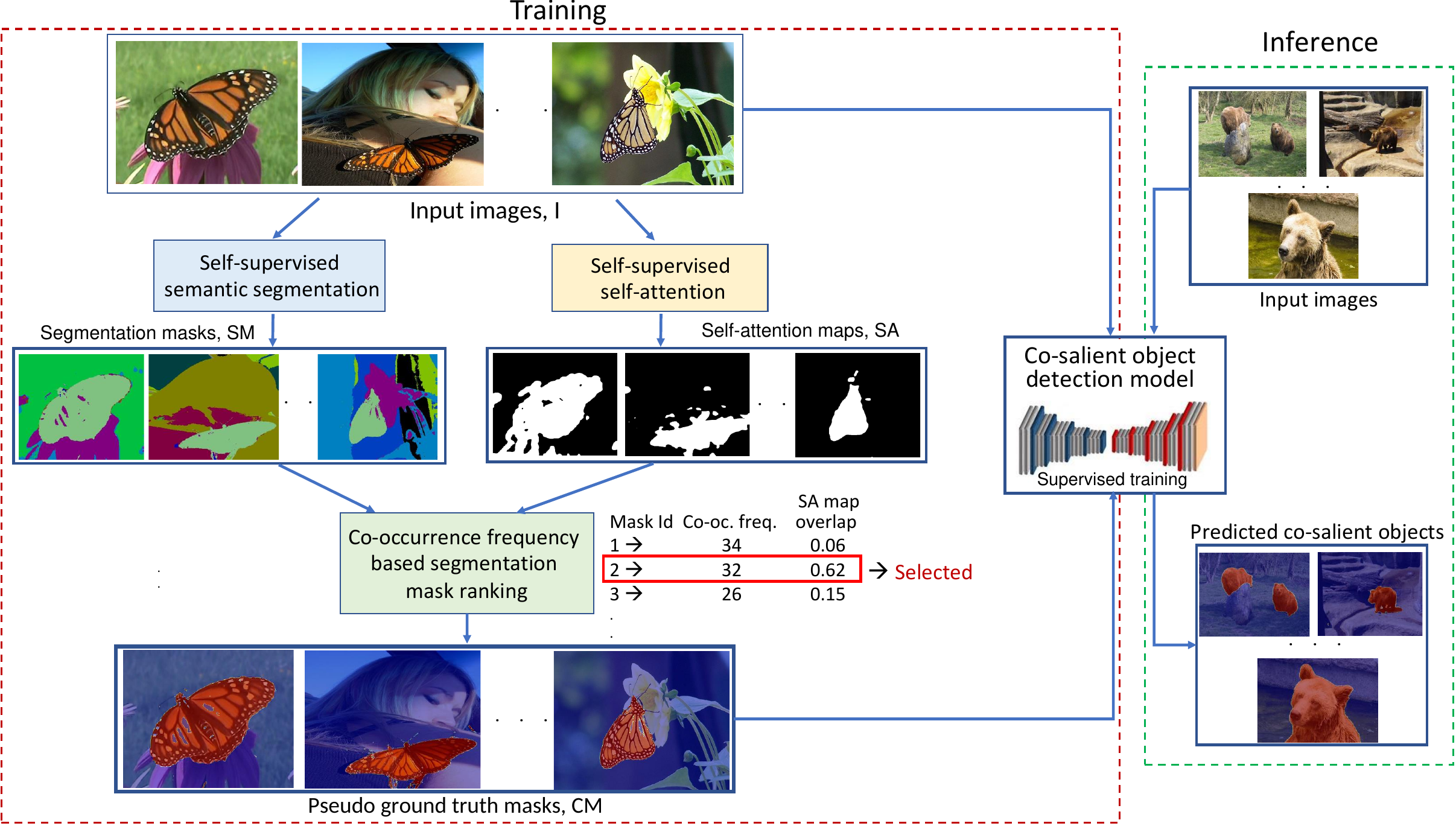}
\caption {The proposed unsupervised CoSOD model, US-CoSOD. We first obtain  unsupervised semantic segmentation maps from STEGO \cite{hamilton2022unsupervised} and self-attention maps from DINO \cite{caron2021emerging}. The  unlabeled categories (from STEGO) in the image group are sorted based on their co-occurrence frequencies and the final category is chosen based on the overlap between the STEGO and DINO masks per image. The segmentation  mask of the selected category is considered the pseudo ground truth for training an off-the-shelf supervised CoSOD model.}
\label{fig:unsup_figure}
\vspace{-1.3mm}
\end{figure*}

\textbf{Semi-supervised segmentation:} While consistency-based methods enforce the predictions of unlabeled samples to be consistent under different perturbations \cite{chen2021semi,lai2021semi,mittal2019semi}, pseudo-label based methods \cite{lee2021anti,wang2022semi,liu2021semi,kwon2022semi,mendel2020semi,ke2020guided},  incorporate unlabeled data into training with high-quality pseudo labels. Some of these methods \cite{kwon2022semi,mendel2020semi,ke2020guided, wang2022semi} apply pixel-level error correction mechanisms on the generated pseudo-labels (\eg~using an auxillary decoder or  employing a flaw detector or a discriminator) in order to avoid propagating label noise. Instead, we directly estimate the prediction error probability on an unlabeled image at the global level using a confidence estimation module, trained to estimate prediction confidence (in terms of the segmentation accuracy) based on the labeled set to block error propagation. Wang et al.~\cite{6751158} proposed the first semi-supervised co-segmentation model by optimizing an energy function consisting of inter- and intra-image distances for an image group. While Zheng et al.~\cite{semisupcosod} proposed the first semi-supervised CoSOD framework based on graph structure optimization, their model accuracy is low due to the use of hand-crafted features and their model has not been evaluated under sufficiently low labeled data. Some semi-supervised SOD models have also been proposed \cite{zhang2020few,liu2021semi,lv2022semi}. SAL \cite{lv2022semi} used active learning to gradually expand a small labeled set to include samples on which predictions are inaccurate. GWSCoSal \cite{qian2022co} introduced a weakly supervised learning induced CoSOD model using  group class activation maps.\\
\vspace{-0.2mm}
Existing unsupervised and semi-supervised approaches suffer from limited performance because they use: (1) hand-crafted features, and (2) smaller unlabeled datasets. Our study fills this gap by introducing an unsupervised and a semi-supervised CoSOD method, both of which uses large-scale unlabeled data to 
significantly improve performance.

\vspace{-1.1mm}
\section{Methodology}
\vspace{-1mm}
\label{sec:method}
 Given a group of $N$ images $I = \{I_1, I_2, ..., I_n\}$ containing co-occurring salient objects of a certain class, CoSOD
aims to detect them simultaneously and output their co-salient object segmentation masks. For unsupervised CoSOD, the goal is to predict the co-salient segmentations $\{\hat{y}_i\}_{i=1}^n$ without using any labeled data. For semi-supervised CoSOD, given a labeled set $\mathcal{D}_l = \{(x^l_i,y^l_i)\}_{i=1}^{N_l}$ and a much
larger unlabeled set $\mathcal{D}_u = \{(x^u_i)\}_{i=1}^{N_u}$, we aim to train a CoSOD model by efficiently utilizing both the limited labeled data and a large amount of unlabeled data.

\begin{algorithm}
\caption{Pseudo co-saliency mask generation}
 \label{alg:algorithm1}
 \begin{algorithmic}[1]
 \renewcommand{\algorithmicrequire}{\textbf{Input: }}
 \renewcommand{\algorithmicensure}{\textbf{Output:}}
 \REQUIRE Image group I = \{$I_1,I_2,..,I_n$\}
 \ENSURE  CoSOD masks $CM$ = \{$CM_1,CM_2,..,CM_n$\}
\\
\rule{\linewidth}{0.4pt}  
  \STATE Obtain self-attention (SA) maps, $SA = \{ SA_i\}_{i=1}^n$ from DINO \cite{caron2021emerging}.
  
  \STATE Apply Otsu thresholding on the SA maps to obtain binary segmentation maps, $DM = \{DM_i\}_{i=1}^n$.
  
  \STATE Obtain the unsupervised single-image semantic segmentation maps, $SM_i^c$ for each image $i$ and discovered unlabeled category $c$ from STEGO \cite{hamilton2022unsupervised}.
  
  \STATE Compute the frequency, $f^{c}$ of each semantic unlabeled category c from STEGO in the image group, $I$.

  \FOR {$i = 1$ to $n$}
    \STATE  $C_i = \{c^1,c^2,...,c^m\}$ is the set of discovered unlabeled categories in the image $I_i$.
    \STATE Sort the categories in $C_i$ by their frequency $f^c$ in the descending order and select the top-$K$ frequent unlabeled categories, $U = \{u^1,u^2,..,u^K\}$.
    \FOR {$j = 1$ to $K$}
      
      \STATE For category $u^j$,  compute overlap score:\\ $O^{u^j}_i = Ar({SM^{u^j}_i} \cap DM_i)/Ar(I_i)$, the overlapped area between the STEGO mask $SM^{u^j}_i$ for category $u^j$ and the DINO SA map $DM_i$ divided by the total image area.

      \ENDFOR \\
      Co-salient object mask, $CM_i = SM^{c^{coso}_i}_i$ is the STEGO mask of the class, $c^{coso}_i$ that maximizes the overlap score $O_i$ i.e. $c^{coso}_i = \arg \max_c O^c_i$
  \ENDFOR
 \RETURN $CM$ 
 \end{algorithmic} 
 \end{algorithm}

\vspace{-1.3mm}
\subsection{Unsupervised co-salient object detection}
Here, we describe our unsupervised CoSOD model (US-CoSOD) that effectively leverages the frequency statistics of self-supervised single-image semantic segmentations.  

Fig.~\ref{fig:unsup_figure} depicts our unsupervised pipeline for CoSOD. We first compute the pseudo co-saliency masks based on the single image segmentation masks and the self-attention masks, which are then used to train a fully supervised CoSOD model. Trained with the self-distillation loss \cite{hinton2015distilling}, the attention maps associated with the class token from the last layer of DINO  \cite{caron2021emerging} have been shown to highlight salient foreground regions \cite{caron2021emerging,wang2022tokencut,yin2022transfgu}. Motivated by this observation, we consider the averaged attention map (across all attention heads) from DINO as the foreground object segmentation. Also, to detect the  co-occurring objects, we use the semantic segmentation masks from a recent self-supervised single-image semantic segmentation model, STEGO  \cite{hamilton2022unsupervised}. This model shows that feature correspondences across images form strong signals for unsupervised semantic segmentation. These correspondences are used to create pixel-wise embeddings, which yield high quality semantic segmentation maps upon clustering. We consider these co-occurring semantic clusters across the image group as unlabeled categories and leverage them to find the co-occurring objects. 

We detail our unsupervised pseudo co-saliency  mask generation in Algorithm \ref{alg:algorithm1}. First, we average the self-attention maps from the $n_{h}$ DINO attention heads to obtain the averaged  self-attention map $SA_i$ for an image $I_i$ as: $SA_i = \frac{1}{n_{h}}\sum_{j=1}^{n_{h}} AM_i^j$, where $AM_i^j$ is the attention map from the DINO attention head $j$ for the image $I_i$. Map $SA_i$ is normalized by min-max normalization. We then find the co-occurrence frequency $f^c$ of the discovered categories across all images in the group $I$. Next, for each image $I_i$, we compute the top-$K$ frequent STEGO categories and finally select a single unlabeled category $c^{coso}_i$ per image based on the  overlap score $O^c_i$ between the STEGO and the DINO masks. We then consider the STEGO mask $SM^{c^{coso}_i}_i$ corresponding to the category $c^{coso}_i$  as the pseudo ground truth mask for $I_i$. This filtering step ensures that the selected segmentation corresponds to the most common yet salient object in the group, therefore preventing co-occurring backgrounds from being considered as the pseudo masks. 

Thus we obtain the pseudo co-salient object masks $CM_{train}$ for all groups $I_{train}$ in our training set and train a CoSOD model \cite{yu2022democracy} in a supervised manner using $I_{train}$ as input and the corresponding pseudo segmentation masks $CM_{train}$ as the training labels. The training loss, $\mathcal{L}^{unsp}$ is:
\vspace{-2mm}
\begin{equation}
    \mathcal{L}^{unsp} = \frac{1}{|B|} \sum_{i=1}^{|B|} l_{iou}(f^{unsp}(x_i,\theta),CM_i) + \lambda_{sc} l_{sc}
\end{equation}
where, $l_{iou}$ is the IoU loss \cite{zhang2020coadnet,qin2019basnet} between the predicted segmentation map, $f^{unsp}(x_i,\theta)$ and the ground truth segmentation $CM_i$, $B$ is the training batch, and $x_i$ is the input image. $l_{sc}$ is the self-contrastive loss as outlined in \cite{yu2022democracy}. 

At inference, we use the trained $f^{unsp}$ model to detect co-salient objects in the test image groups, $I_{test}$. Note that the self-supervised component models (STEGO and DINO) are only used during the training of our US-CoSOD model (to generate the pseudo-labels) and not during inference.

\begin{figure*}
\centering
\includegraphics[width = 0.876\textwidth]{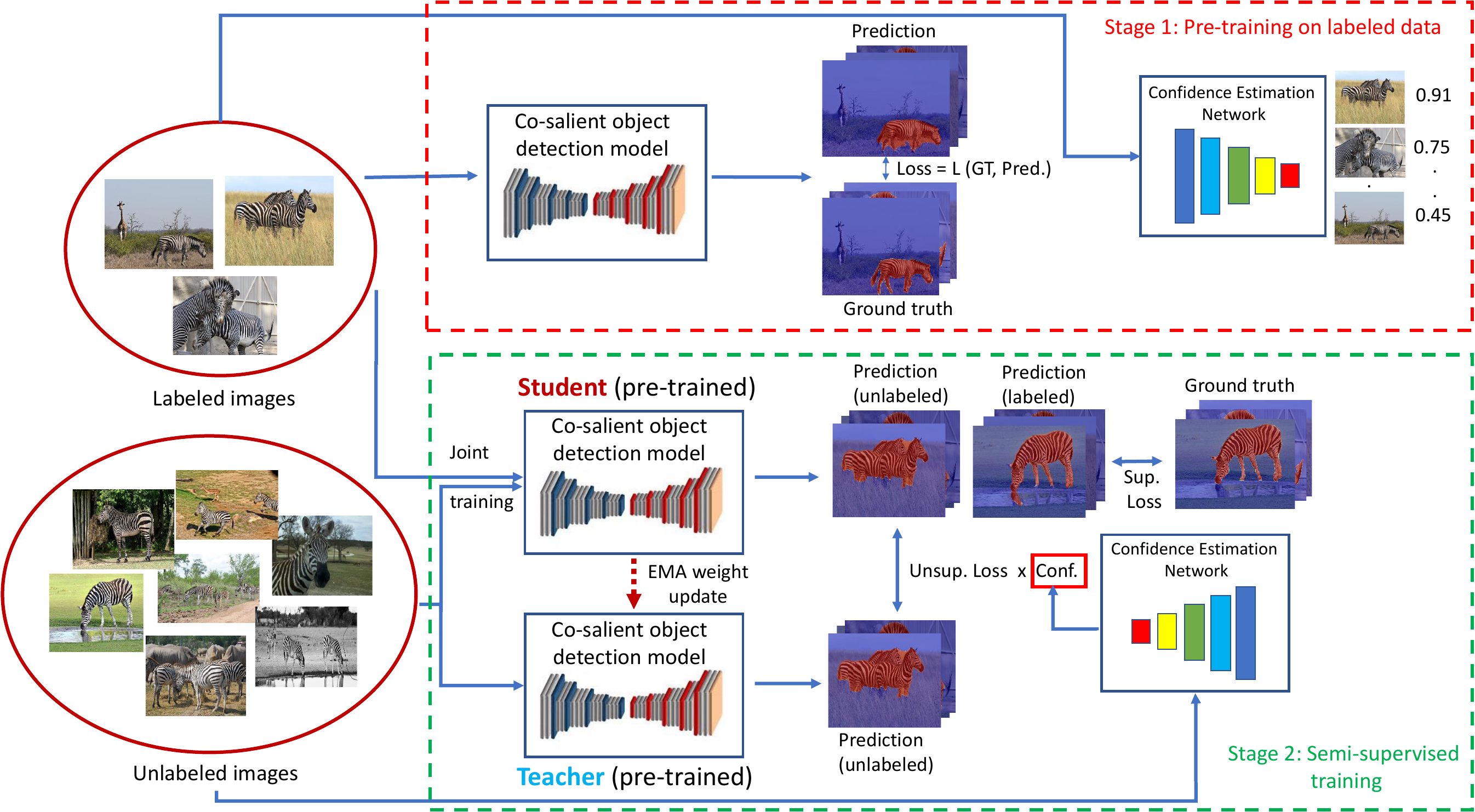}
\caption {Proposed SS-CoSOD model for semi-supervised co-salient object detection. In the first stage, we pre-train a CoSOD model on the labeled set while also training our Confidence Estimation  Network (CEN) on the same set. In stage 2, we employ a student-teacher model (initialized with the model from stage 1) for semi-supervised learning. The labeled and unlabeled data are jointly passed through the student during training. We weight the unsupervised loss by the confidence score estimated by the CEN module.}
\label{fig:semisup_figure}
\vspace{-1.3mm}
\end{figure*}

\vspace{-0.6mm}
\subsection{Semi-supervised co-salient object detection}
While the unsupervised model does not require any labeled data for training, such a model is often not at par with supervised models in terms of prediction accuracy. Therefore, we develop a semi-supervised approach for CoSOD (SS-CoSOD), which can effectively leverage a large amount of unlabeled data with effective prediction confidence estimation. Fig.~\ref{fig:semisup_figure} depicts our semi-supervised pipeline. 

In stage 1, we supervisedly pre-train a CoSOD model $f_{PT}$ on the labeled set and then train our Confidence Estimation  Network (CEN) on the same set. In stage 2, we employ a  typical self-training framework with two models of the same architecture, namely student (model $f_S$) and teacher (model $f_T$) networks that are initialized as $f_{PT}$ in stage 1. The student model’s weights $\theta_s$ are updated via  backpropagation while the teacher model’s weights $\theta_t$ are updated with the  exponential moving average (EMA) scheme  \cite{tarvainen2017mean}, \ie, $\theta_t = \lambda_d \theta_{t-1} + (1 - \lambda_d) \theta_s$, where $\lambda_d$ is the EMA decay factor (set to 0.95). At each training step, we sample $B_l$ labeled images  and $B_u$ unlabeled images (maximum value of $|B_l|$ and $|B_u|$ being 16, following \cite{yu2022democracy}). Next, we combine each batch of labeled images, $B_l$ and unlabeled images, $B_u$ into a single volume $B_{l+u}$ before passing the combined volume through the student network. The training label for this volume is the combination of the ground truth labeled mask with the teacher model prediction as: $y^{(l+u)} = [y^l,f_T(x^u_j,\theta_T)]$. This step is different from other semi-supervised   approaches, \eg, U2PL \cite{wang2022semi} where the labeled and the unlabeled sets are passed through the student model in two separate passes. We leverage the benefit of learning cross-pixel similarities across all images in the image group to effectively propagate object co-saliency information from the labeled images to the unlabeled images. 

A supervised loss, $\mathcal{L}_{s}$ is computed for the labeled set and an unsupervised loss, $\mathcal{L}_{u}$ is computed for the unlabeled samples. For every labeled image, our goal is to minimize the supervised IoU loss as: 
\vspace{-1.2mm}
\begin{equation}
    \mathcal{L}_{s} = \frac{1}{|B_l|} \sum_{\substack{(x^l_i,y^l_i)\in B_l,\\x^u_j\in B_u}} l_{iou}(f_S^l([x^l_i,x^u_j],\theta_S),y_i^l) + \lambda_{sc} l_{sc}
\end{equation}
where $l_{iou}$ is IoU loss and $l_{sc}$ is self-contrastive (SC) loss. $\lambda_{sc}$ is set as 0.1 (following \cite{yu2022democracy}). The SC Loss, $l_{sc}$ is computed as: $l_{sc} = -\log(cos_c + \epsilon) - \log(1-\cos_b + \epsilon)$ with $cos_c = cos(proto^{(l+u)},proto_c^{(l+u)})$ and $cos_b = cos(proto^{(l+u)},proto_b^{(l+u)})$, 
where $proto$ is the prototype generated by the original inputs, $proto_c$ is the co-salient prototype generated by the foreground regions, and $proto_b$ is the background prototype generated from the background regions in the image.  $cos$ is the cosine-similarity function and $\epsilon$ is a small positive constant to avoid overflow.  

\begin{table*}[t]
\scriptsize
\centering
\caption{
Performance comparison of unsupervised models for CoSOD. Our US-CoSOD-ImgNet150 model achieves the best result.
}
\label{tab:classic}
\setlength{\tabcolsep}{3pt}
\vspace{1pt}
\resizebox{16.5cm}{!}{
\begin{tabular}{l|llll|llll|llll}
\toprule
 & \multicolumn{4}{c}{CoCA} & \multicolumn{4}{c}{Cosal2015} &
 \multicolumn{4}{c}{CoSOD3k}\\
Method &
MAE$\downarrow$ & $F_{\beta}^{max}\uparrow$ & $E_{\phi}^{max}\uparrow$ & $S_{\alpha}\uparrow$ & MAE$\downarrow$ & $F_{\beta}^{max}\uparrow$ & $E_{\phi}^{max}\uparrow$ & $S_{\alpha}\uparrow$ &
MAE$\downarrow$ & $F_{\beta}^{max}\uparrow$ & $E_{\phi}^{max}\uparrow$ & $S_{\alpha}\uparrow$\\

\midrule
 DINO (DI) \cite{caron2021emerging} (ICCV 2021) &
0.214 & 0.372 & 0.572 & 0.540 & 0.154 & 0.659 & 0.753 & 0.688 & 0.146 & 0.624 & 0.749 & 0.679\\

 STEGO (ST)  \cite{hamilton2022unsupervised} (ICLR 2022) &
0.235 &  0.353 & 0.555 & 0.523 & 0.164 & 0.618 & 0.717 & 0.676 & 0.204 & 0.543 & 0.660 & 0.615\\

TokenCut \cite{wang2022self} (CVPR 2022) &
0.167 & 0.467 & 0.704 & 0.627 & 
0.139 & 0.805 & 0.857 & 0.793 & 
0.151 & 0.720 & 0.811 & 0.744\\
 
 DVFDVD \cite{amir2021deep} (ECCVW 2022) &
0.223 &  0.422 & 0.592 & 0.581 & 0.092 & 0.777 & 0.842 & 0.809 & 0.104 & 0.722 & 0.819 & 0.773\\

 SegSwap \cite{shen2022learning} (CVPRW 2022) &
0.165 &  0.422 & 0.666 & 0.567 & 0.178 & 0.618 & 0.720 & 0.632 & 0.177 & 0.560 & 0.705 & 0.608\\

Ours (DI+ST) &
0.165 & 0.461 & 0.676 & 0.610 & 
0.112 & 0.760 & 0.823 & 0.767 &
0.124 & 0.684 & 0.793 & 0.724\\

Ours (US-CoSOD-COCO9213)  &
0.140 & 0.498 & 0.702 & 0.641 & 0.090 & 0.792 & 0.852 & 0.806 & 0.095 & 0.735 & 0.832 & 0.772\\

Ours (US-CoSOD-ImgNet150) &
\textbf{0.116} & \textbf{0.546} & \textbf{0.743} & \textbf{0.672} & \textbf{0.070} & \textbf{0.845} & \textbf{0.886} & \textbf{0.840} & \textbf{0.076} & \textbf{0.779} & \textbf{0.861} & \textbf{0.801}\\

Ours (US-CoSOD-ImgNet450) &
0.127 & 0.543 & 0.726 & 0.666 & 0.071 & 0.844 & 0.884 & 0.842 & 0.079 & 0.775 & 0.854 & 0.800 \\

\bottomrule
\end{tabular}
}
\label{tab:unsup_cosod}
\vspace{-3pt}
\end{table*}

We pass the unlabeled batch, $B_u$, through the teacher network and compute the unsupervised loss between the predictions of the teacher and the student networks. The unsupervised loss is weighted using the confidence scores predicted by the CEN module for each unlabeled image as: 
\vspace{-3.6mm}
\begin{equation}
\label{eq:conf_eq}
\mathcal{L}_{u} = \frac{1}{|B_u|} \sum_{\substack{x^l_i\in B_l, \\ x^u_j\in B_u}} \tilde{g}({x^u_i},\theta_C^l) l_{iou}(f_S^u([x^l_i,x^u_j],\theta_S),f_T(x^u_j,\theta_T))
\end{equation}
where $\tilde{g}({x^u_j},\theta_C^l) = \frac{g({x^u_j},\theta_C^l)}{\sum_{j=1}^{|B_u|} g({x^u_j},\theta_C^l)}$
is the confidence weight estimated by CEN, $g$ for the unlabeled sample $x^u_i$, parameterized by $\theta_C^l$ that is learned from the labeled batch $B_l$. We observed that the normalized confidence weight is crucial for model convergence.
Our objective is to minimize the overall loss, $\mathcal{L} = \mathcal{L}_{s} + \lambda_u \mathcal{L}_{u}$, where $\mathcal{L}_{s}$ and $\mathcal{L}_{u}$ represent supervised loss on the labeled 
set and unsupervised loss on the unlabeled set respectively. $\lambda_u$ is set as 1 \cite{yu2022democracy}.
\\

\vspace{-2.5mm}
\noindent \textbf{Confidence Estimation Network (CEN):} The CEN model is trained to estimate the reliability score of the model prediction. To train this model, we use the labeled image set $S_l$ as the input and the corresponding segmentation F-measure \cite{achanta2009frequency} scores of the pre-trained model $f_{PT}$ predictions in stage 1 as the ground truth. We use a ResNet50 backbone~\cite{he2016deep} trained using the DINO method because of its ability to well segment the discriminative image regions. The model is trained by fine-tuning the pre-trained ResNet backbone and an $fc(2048,1)$ layer using the MSE loss as:
\begin{equation}
     \mathcal{L}_{C} = \frac{1}{|S_l|} \sum_{\substack{(x^l_i,y^l_i)\in S_l}}(CEN(x^l_i,\theta^l_C)-F_\beta(f_{PT}(x^l_i,\theta_S),y^l_i))^2 
\end{equation}
where $f_S$ is the student model parameterized by $\theta_S$, $F_\beta(p,q)$ is  the $F_\beta$-metric computed between two segmentation maps $p$ and $q$, and $S_l$ is the labeled set. After training, CEN provides a reliability estimate of model prediction solely based on the image contents (Fig. 4 in  supplement). 

\vspace{-2mm}
\section{Experimental Results}
\label{sec:results}
\vspace{-1.0mm}
\subsection{Setup}
\noindent \textbf{Datasets and evaluation metrics:} We used labeled data from COCO9213 \cite{wang2019robust}, a subset of the COCO dataset \cite{lin2014microsoft} that  contains 9,213 images selected from 65 groups, to train our semi-supervised model. We additionally constructed a dataset of 150K unlabeled images by selecting 150 images per class from the  training subset of ImageNet-1K  \cite{krizhevsky2017imagenet,deng2009imagenet} to train our unsupervised and semi-supervised models. We evaluate our methods on three popular CoSOD benchmarks: CoCA \cite{zhang2020gradient}, Cosal2015 \cite{zhang2016detection} and
CoSOD3k \cite{fan2020taking}. CoCA and CoSOD3k are 
challenging real-world co-saliency evaluation datasets, containing
multiple co-salient objects in some images, large appearance and scale variations, and complex backgrounds. 
Cosal2015 is a widely used  dataset for CoSOD evaluation. Our evaluation metrics include the  
Mean Absolute Error (MAE$\downarrow$) \cite{cheng2013efficient}, maximum F-measure ($F_{\beta}^{max}\uparrow$) \cite{achanta2009frequency}, maximum E-measure ($E_{\phi}^{max}\uparrow$) \cite{fan2018enhanced}, and S-measure ($S_{\alpha}\uparrow$) \cite{fan2017structure}.

\noindent \textbf{Implementation details:}
We used DCFM \cite{yu2022democracy} as the backbone model for all experiments. While the student network in SS-CoSOD consists of the Democratic Prototype Generation Module (DPG) and the Self-Contrastive Learning Module (SCL) from DCFM, we removed SCL in the teacher network because this network is not updated via backpropagation \cite{yu2022democracy}. We used the Adam optimizer for training. The total training time is 5 hours for US-CoSOD  and 8 hours for SS-CoSOD using ImageNet-1K. The inference time is 84.4 fps. More details in the supplementary.

\subsection{Quantitative evaluation}

\begin{table*}[t]
\scriptsize
\centering
\caption{
Performance comparison of the different versions of our unsupervised and semi-supervised models. In column 1, we indicate the fraction of labeled data for training, followed by the actual number of images. See supplementary for the results with 1/8 labeled data.
}
\label{tab:semisup_cosod}
\setlength{\tabcolsep}{3pt}
\vspace{1pt}
\resizebox{16.7cm}{!}{%
\begin{tabular}{l|l|llll|llll|llll}
\toprule
 & & \multicolumn{4}{c}{CoCA} & \multicolumn{4}{c}{Cosal2015} &
 \multicolumn{4}{c}{CoSOD3k}\\
Labeled data & Method &
MAE$\downarrow$ & $F_{\beta}^{max}\uparrow$ & $E_{\phi}^{max}\uparrow$ & $S_{\alpha}\uparrow$ &
MAE$\downarrow$ & $F_{\beta}^{max}\uparrow$ & $E_{\phi}^{max}\uparrow$ & $S_{\alpha}\uparrow$ &
MAE$\downarrow$ & $F_{\beta}^{max}\uparrow$ & $E_{\phi}^{max}\uparrow$ & $S_{\alpha}\uparrow$ \\

\toprule
 &  US-CoSOD & 0.108 & 0.557 & 0.754 & 0.683 & 
0.068 & 0.854 & 0.888 & 0.846 & 
0.076 & 0.783 & 0.857 & 0.801\\

  & SS-CoSOD-DJ (w/o CEN) &
0.107 & 0.485 & 0.728 & 0.635
& 0.094 & 0.771 & 0.834 & 0.771 &
0.089 & 0.709 & 0.817 & 0.742\\

1/16 (576)  & SS-CoSOD-DJ (w/ CEN)  &
0.115 & 0.488 & 0.730 & 0.639 &  
0.086 & 0.782 & 0.847 & 0.787 &
0.086 & 0.717 & 0.828 & 0.755 \\

  & SS-CoSOD  &
0.113 & 0.492 & 0.733 & 0.641 &
0.085 & 0.788 & 0.850 & 0.792 & 
0.084 & 0.721 & 0.830 & 0.758 \\

  & US-CoSOD+SS-CoSOD &
0.111 & 0.554 & 0.751 & 0.681 &
\textbf{0.066} & \textbf{0.855} & \textbf{0.890} & \textbf{0.849} &
0.075 & 0.783 & 0.858 & \textbf{0.803}\\

  & SS-CoSOD with ImgNet &
\textbf{0.098} & \textbf{0.562} & \textbf{0.757} & \textbf{0.684} &
0.072 & 0.837 & 0.880 & 0.828 &
\textbf{0.068} & \textbf{0.784} & \textbf{0.865} & 0.800\\

\toprule
 &  US-CoSOD & 0.109 & 0.569 & 0.758 & 0.685 &
0.069 & 0.855 & 0.888 & 0.844 &
0.077 & 0.783 & 0.854 & 0.797\\
 
  & SS-CoSOD-DJ (w/o CEN) &
0.097 & 0.552 & 0.763 & 0.678  &
0.076 & 0.828 & 0.874 & 0.818  &
0.075 & 0.776 & 0.859 & 0.790 \\

 1/4 (2303)  & SS-CoSOD-DJ (w/ CEN) &
0.096 & 0.560 & 0.764 & 0.685  &
0.069 & 0.839 & 0.885 & 0.831 &
0.069 & 0.784 & 0.867 & 0.802  \\

  & SS-CoSOD  &
0.097 & 0.562 & 0.765 & 0.686  &
0.068 & 0.841 & 0.886 & 0.833 &
0.068 & 0.785 & 0.868 & 0.803  \\

  & US-CoSOD+SS-CoSOD &
0.107 & 0.566 & 0.757 & 0.686  &
0.066 & \textbf{0.858} & 0.891 & \textbf{0.848} &
0.073 & 0.787 & 0.859 & 0.803\\

  & SS-CoSOD  with ImgNet  &
\textbf{0.091} & \textbf{0.581} & \textbf{0.772} &  \textbf{0.698} &
\textbf{0.066} & 0.851 & \textbf{0.891} & 0.841  &
\textbf{0.064} & \textbf{0.799} & \textbf{0.875} & \textbf{0.812} \\

\toprule
  &  US-CoSOD & 0.105 & 0.569 & 0.760 & 0.688 &
0.068 & 0.856 & 0.889 & 0.843 &
0.074 & 0.793 & 0.862 & 0.804\\
 
   & SS-CoSOD-DJ (w/o CEN) &
0.092 & 0.572 & 0.771 & 0.694 &
0.068 & 0.846 & 0.885 & 0.834  &
0.071 & 0.791 & 0.865 & 0.802\\

1/2 (4607)  & SS-CoSOD-DJ (w/ CEN) &
0.090 & 0.578 & 0.772 & 0.699 &
0.062 & 0.851 & 0.892 & 0.843 &
0.067 & 0.795 & 0.870 & 0.810\\

  & SS-CoSOD  &
0.088 & 0.582 & 0.773 & 0.700  &
0.062 & 0.854 & 0.892 & 0.843  &
0.066 & 0.797 & 0.872 & 0.809 \\

  & US-CoSOD+SS-CoSOD &
0.110 & 0.563 & 0.755 & 0.686 &
0.064 & 0.858 & 0.894 & 0.850 &
0.072 & 0.794 & 0.866 & 0.810 \\

   & SS-CoSOD with ImgNet  &
\textbf{0.088} & \textbf{0.590} & \textbf{0.775} & \textbf{0.705} &
\textbf{0.062} & \textbf{0.861} & \textbf{0.896} & \textbf{0.850} &
\textbf{0.063} & \textbf{0.804} & \textbf{0.876} & \textbf{0.817}\\

\midrule
Full (9213) & US-CoSOD & 
0.102 & 0.573 & 0.764 & 0.692 &
0.068 & 0.860 & 0.890 & 0.845 &
0.077 & 0.791 & 0.856 & 0.799\\

 & SS-CoSOD with ImgNet & \textbf{0.091} & \textbf{0.591} & \textbf{0.778} & \textbf{0.707} & \textbf{0.061} & \textbf{0.865} & \textbf{0.901} & \textbf{0.852} & \textbf{0.062} & \textbf{0.809} & \textbf{0.882} & \textbf{0.821}\\
 
\midrule
\end{tabular}
}
\vspace{-5pt}
\end{table*}

In Table \ref{tab:unsup_cosod}, we quantitatively evaluate the predictions obtained using seven different unsupervised baseline models: DINO self-attention mask (DI) \cite{caron2021emerging}, the most frequently co-occurring semantic unlabeled category mask from STEGO (ST) \cite{hamilton2022unsupervised}, the pseudo co-saliency mask (DI+ST), predictions from TokenCut \cite{wang2022self}, DVFDVD \cite{amir2021deep}, SegSwap \cite{shen2022learning}, and predicted masks from our US-CoSOD model. We  evaluated SegSwap \cite{shen2022learning} (that only predicts pairwise co-segmentations) by averaging the  predicted co-segmentations over all image pairs in an image group i.e. between one image and every other image in the group.  
Table \ref{tab:unsup_cosod} shows that our US-CoSOD model trained on the  150K images from ImageNet (150 images per class) achieves the best performance. However, increasing the number of images to 450 per class slightly reduced the  model accuracy. This could be attributed  to the fact that adding more unlabeled images to the training set may lead to erroneous training due to the noisy pseudo ground truth masks generated by DI+ST (using which US-CoSOD is trained). 

In Table \ref{tab:semisup_cosod}, we compare the performance of different versions of our SS-CoSOD models using different proportions of labeled data. The ``US-CoSOD'' prefix indicates that the model is initialized with the pre-trained US-CoSOD model. ``DJ'' indicates that the model is trained with labeled and unlabeled images passed `disjointly' to the student model without any cross-region interaction between the two sets. The SS-CoSOD model is trained semi-supervisedly using only images from the COCO9213 dataset. Finally, the ``SS-CoSOD with ImgNet'' version utilizes the extra 150 images per 1K ImageNet classes during semi-supervised training. In order to include the unlabeled ImageNet images in our semi-supervised setting, we use CEN to infer the confidence weights for the unlabeled images and then include the samples with predicted score $\geq 0.9$ in the labeled set and the remaining samples in the unlabeled set.

In Table \ref{tab:semisup_cosod}, we observe a consistent improvement in SS-CoSOD performance when the reliability scores from CEN are used to modulate the unsupervised loss. Joint labeled-unlabeled student training further leads to a small but consistent improvement. We further observe a significant improvement in performance when the ImageNet-1K dataset is used for semi-supervised learning, \eg~on the CoCA dataset, we obtain a reduction of 20\% for MAE and gains of 4.79\% for maximum F-measure, 2.64\% for maximum E-measure, and 2.76\% for S-measure compared to SS-CoSOD trained using the COCO9213 dataset on the 1/2 data split.

\begin{figure}
\centering
\includegraphics[width = 0.46\textwidth]{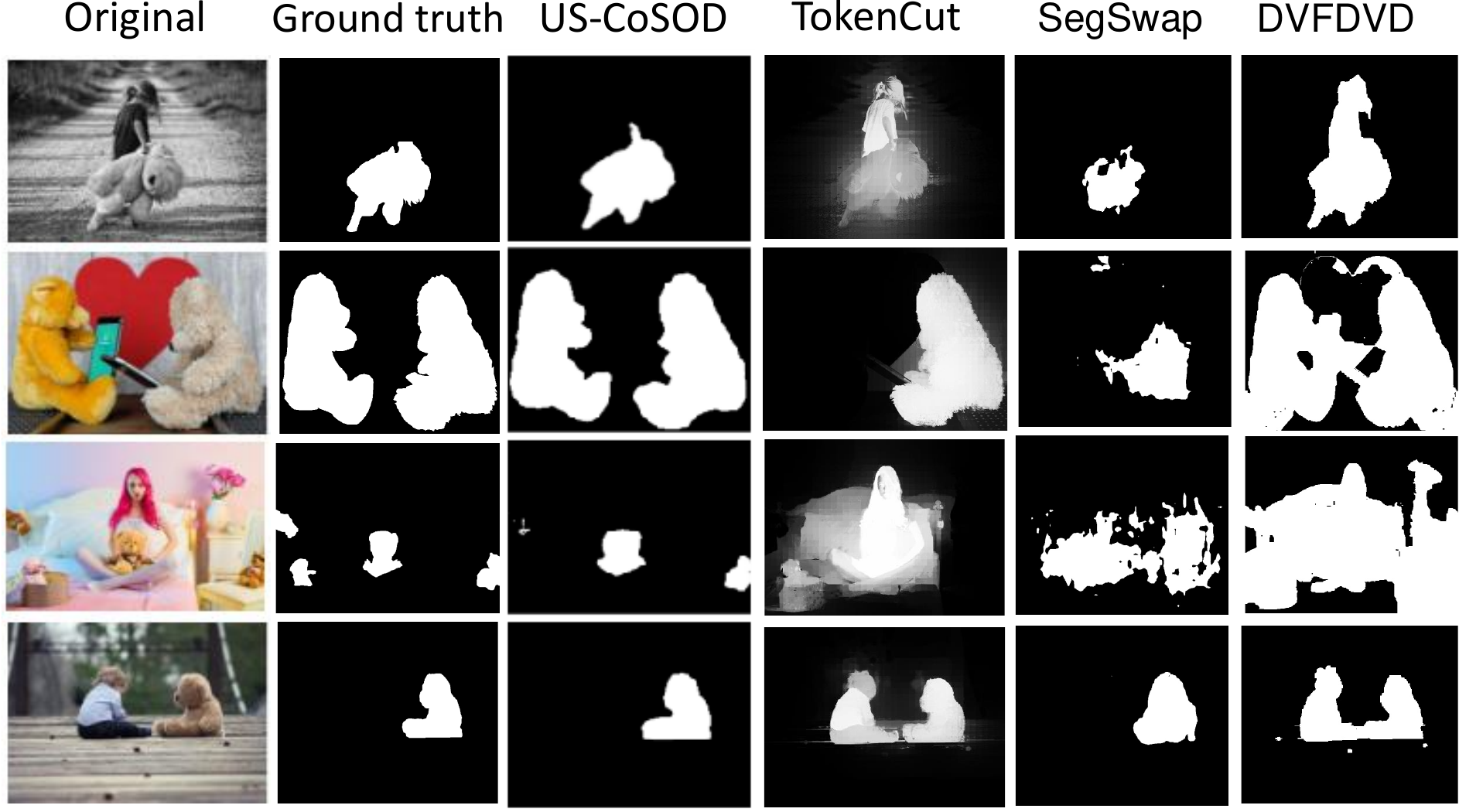}
\caption {Qualitative comparisons of our US-CoSOD model with other baselines on the \textit{teddy bear} image group from CoCA. US-CoSOD produces the most accurate segmentation masks.} 
\label{fig:quals_unsup}
\vspace{-4.3mm}
\end{figure}

\begin{figure*}
\centering
\includegraphics[width = 0.93\textwidth]{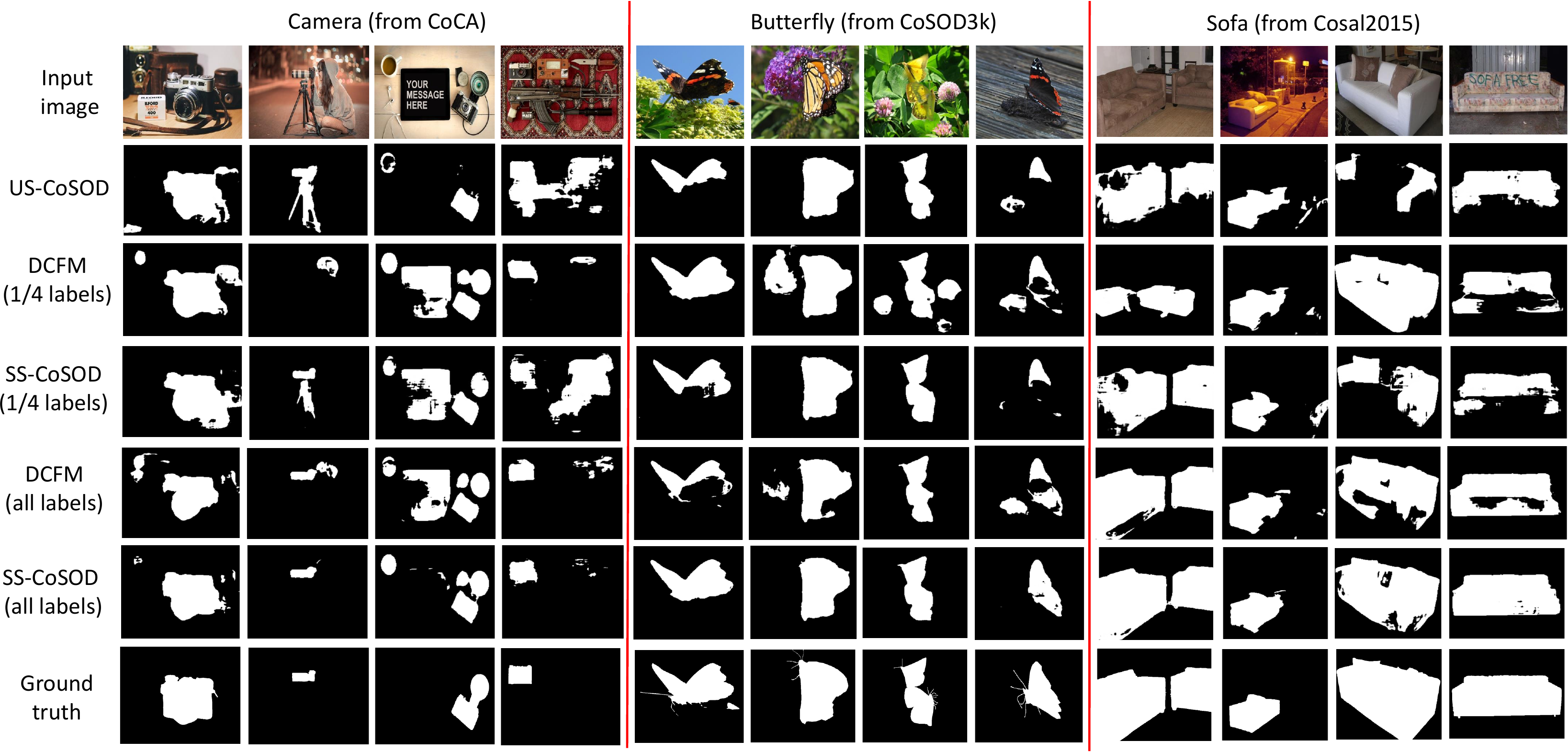}
\caption {Qualitative comparisons of our model with different baselines on three image groups selected each from the CoCA, CoSOD3k, and Cosal2015 datasets. Our SS-CoSOD model (all labels) produces the most accurate segmentation mask compared to the other baselines.} 
\label{fig:qual}
\vspace{-1.3mm}
\end{figure*}

\noindent \textbf{Comparison with existing CoSOD models:} In Table~\ref{tab:unsup_compare}, we compare the performance of  existing unsupervised and semi-supervised CoSOD models with our model using the $F_{\beta}^{max}$-measure and the $S_{\alpha}$-measure metrics. Both of our unsupervised and semi-supervised models  outperform the corresponding state-of-the-art models by a significant margin \eg, on the Cosal2015 dataset, our US-CoSOD model has an 8.8\% F-measure gain over DVFDVD \cite{amir2021deep}, an unsupervised SOTA co-segmentation model and our SS-CoSOD model (using ImageNet-1K) has an 11.81\% F-measure gain over FASS \cite{semisupcosod}, a semi-supervised CoSOD model. 

\vspace{1mm}
\noindent \textbf{Ablation studies:}

\noindent \textbf{Variant model:} For certain image groups (\eg~\textit{key}, \textit{frisbee}, etc.), the co-saliency of the common objects can be lesser than that of other bigger objects solely due to the lesser area, which could impact performance. To test this hypothesis, we investigate a variant model that normalizes the overlap score by STEGO mask area as: 
$O_i^{'j} = \frac{Ar({SM^j_i} \cap DM_i)}{Ar(SM^j_i)}$ (see Algorithm \ref{alg:algorithm1}). US-CoSOD (without area normalization) has more accurate predictions  compared to this variant on all test sets, \eg~US-CoSOD achieves F-measures  0.461, 0.760, and 0.684 against the variant model's  0.410, 0.613, and 0.579 on  CoCA, Cosal2015, and CoSOD3k respectively. See supplementary for more details.\\
\noindent \textbf{Performance on challenging  categories:}  Our US-CoSOD outperforms the pre-trained DINO and DINO+STEGO models by a significant margin on challenging categories (categories over which DINO scored lesser than the average DINO F-measure score over the test dataset). For instance, the average F-measures of DINO, DINO+STEGO, and US-CoSOD are 0.598, 0.654, and 0.738 on the Cosal2015 dataset respectively. See supplementary for more details.
\\
\noindent \textbf{CEN backbone:} We compared the confidence estimation error (Mean Squared Error, MSE) of CEN using different backbone networks on the unlabeled set. DINO (ResNet50) yielded the least MSE across different data splits \eg~MSE using ResNet50, MobileNetV2, ViTB, and ViTS are 0.166, 0.171, 0.176, and 0.177 respectively on the 1/4 labeled data split. See supplementary for details. We attribute the lower accuracy of MobileNetV2 to its lower feature representation power. Also the transformer models DINO (ViTB, ViTS) fail to outperform the convolutional models (\eg~ResNet50) due to the lesser training data  (in different data splits).

\vspace{-1.5mm}
\subsection{Qualitative evaluation}

In Fig.~\ref{fig:quals_unsup}, we qualitatively compare the CoSOD predictions from TokenCut \cite{wang2022self}, SegSwap \cite{shen2022learning}, and DVFDVD \cite{amir2021deep} with our US-CoSOD on the \textit{teddy bear} image group from CoCA. We see that US-CoSOD more accurately detects the teddy bear in the four images compared to the baselines. 

In Fig.~\ref{fig:qual}, we qualitatively compare the CoSOD predictions from different baselines with SS-CoSOD on three image groups, each from the CoCA, CoSOD3k, and Cosal2015 datasets. We see that our US-CoSOD generates reasonable masks and our SS-CoSOD approaches further improve these  predictions while performing better than the fully supervised DCFM trained with limited labels. While DCFM (all labels) produces incomplete segmentations (columns 5, 9, 11, and 12) and overestimates the co-saliency (in columns 2, 3, 5, and 8) in certain image regions, our SS-CoSOD (all labels) model predictions suffer fewer inaccuracies, producing more accurate CoSOD masks.  

\begin{table}[!h]
\scriptsize
\centering
\caption{
Comparison of our models with existing unsupervised and semi-supervised models for CoSOD on Cosal2015.
}
\label{tab:unsup_compare}
\setlength{\tabcolsep}{3pt}
\resizebox{7.4cm}{!}{%
\begin{tabular}{l|l|l|ll}
\toprule
Method & Type & Label \% &
$F_{\beta}^{max}\uparrow$ & $S_{\alpha}\uparrow$ \\

\midrule
CSSCF \cite {jerripothula2016image} (TMM 2016)  & Unsup & - &
 0.682 & 0.671   \\

CoDW \cite {zhang2016detection} (IJCV 2016)  & Unsup & - &
 0.705 & 0.647   \\
 
UCCDGO \cite{hsu2018unsupervised} (ECCV 2018)  & Unsup & - &
0.758 &  0.751  \\

DVFDVD \cite{amir2021deep} (ECCVW 2022)  & Unsup & - &
0.777 &  0.809  \\

SegSwap \cite{shen2022learning} (CVPRW 2022)  & Unsup & - &
0.618 &  0.632  \\

Ours (US-CoSOD) & Unsup & - &
\textbf{0.845} & \textbf{0.840}  \\

\midrule
FASS \cite{semisupcosod} (ACMM 2018)  & Semi-sup & 50\% &
0.770 &   -  \\
Ours (SS-CoSOD w/ ImgNet)  & Semi-sup & 25\% &
0.851 & 0.841\\
Ours (SS-CoSOD w/ ImgNet)  & Semi-sup & 50\% &
\textbf{0.861} & \textbf{0.850}\\

\bottomrule
\end{tabular}}
\vspace{-1mm}
\end{table}

\begin{figure}
\centering
\includegraphics[width = 0.47\textwidth]{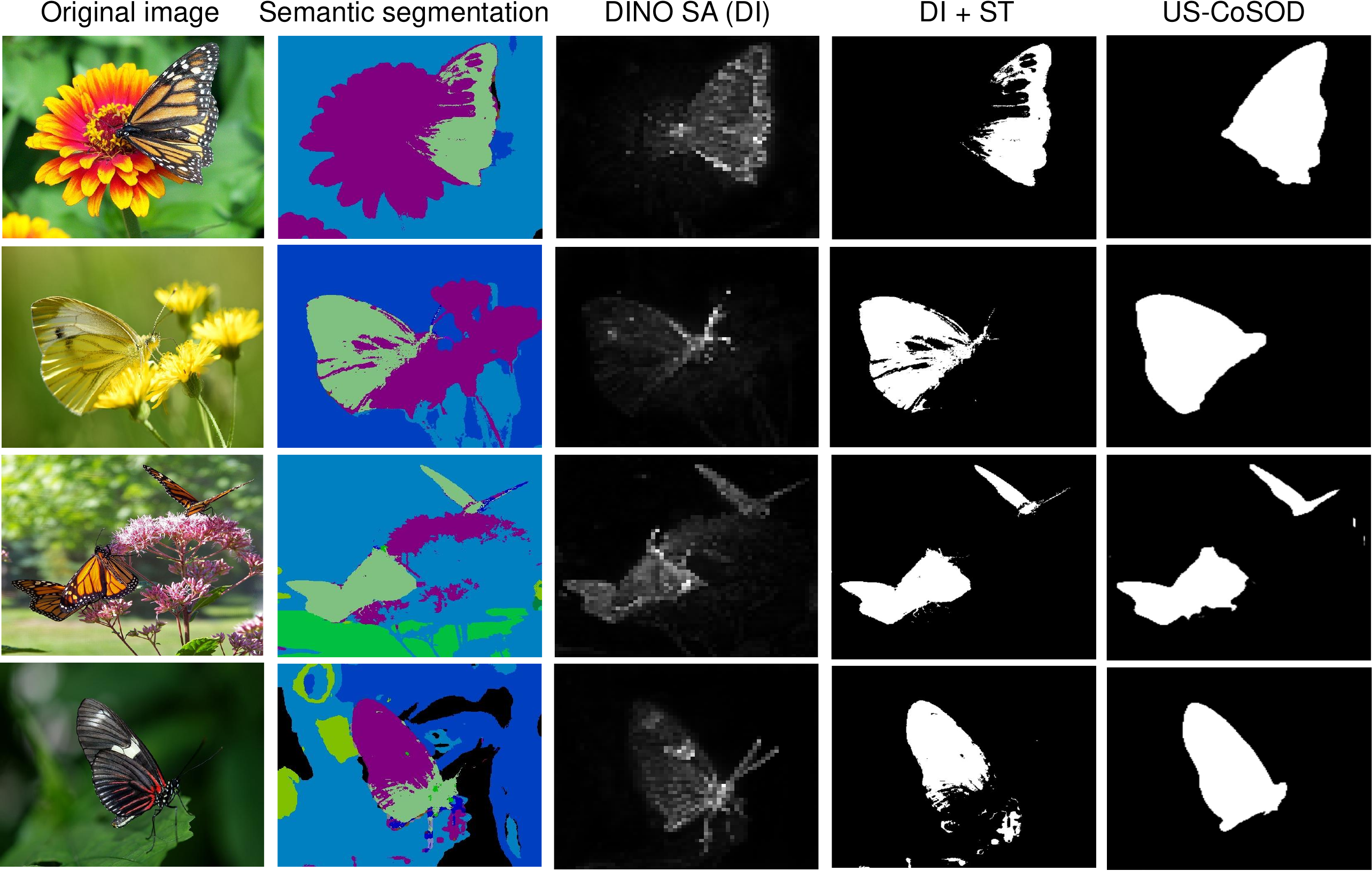}
\caption {DINO self-attention maps and US-CoSOD predictions using DI+ST for training. US-CoSOD produces the best results.}
\label{fig:qual2}
\vspace{-4.2mm}
\end{figure}

Fig.~\ref{fig:qual2} shows that while multiple categories can co-exist in an image  group (e.g. \textit{flowers} and \textit{butterflies}), our DI+ST model extracts the more co-salient butterflies, guided by the DINO SA maps. Also, our US-CoSOD model trained using the pseudo co-saliency masks from DI+ST improves the co-saliency masks (better structural consistency) from DI+ST.

\section{Conclusion}
\label{sec:conclusion}
We presented a novel unsupervised approach  for CoSOD based on the frequency statistics of semantic segmentations, which forms a strong pre-training initialization for a semi-supervised CoSOD model. Our semi-supervised model employs a student-teacher approach with an effective confidence estimation module. We demonstrate that both our unsupervised and semi-supervised CoSOD models can significantly improve prediction performance over a fully-supervised model trained with limited labeled data. As future work, we aim  to improve our unsupervised model by avoiding the use of any off-the-shelf component.

\clearpage

\textbf{\large Supplemental Material:  Unsupervised and semi-supervised co-salient object detection via segmentation frequency statistics}

\setcounter{equation}{0}
\setcounter{figure}{0}
\setcounter{table}{0}
\setcounter{page}{1}
\setcounter{section}{0}
\makeatletter
\renewcommand{\theequation}{S\arabic{equation}}
\renewcommand{\thefigure}{S\arabic{figure}}

\section{Additional qualitative results}
\label{sec:intro}
\subsection{Comparison of CoSOD predictions}

\begin{figure*}
\centering
\includegraphics[width = 17.6cm]{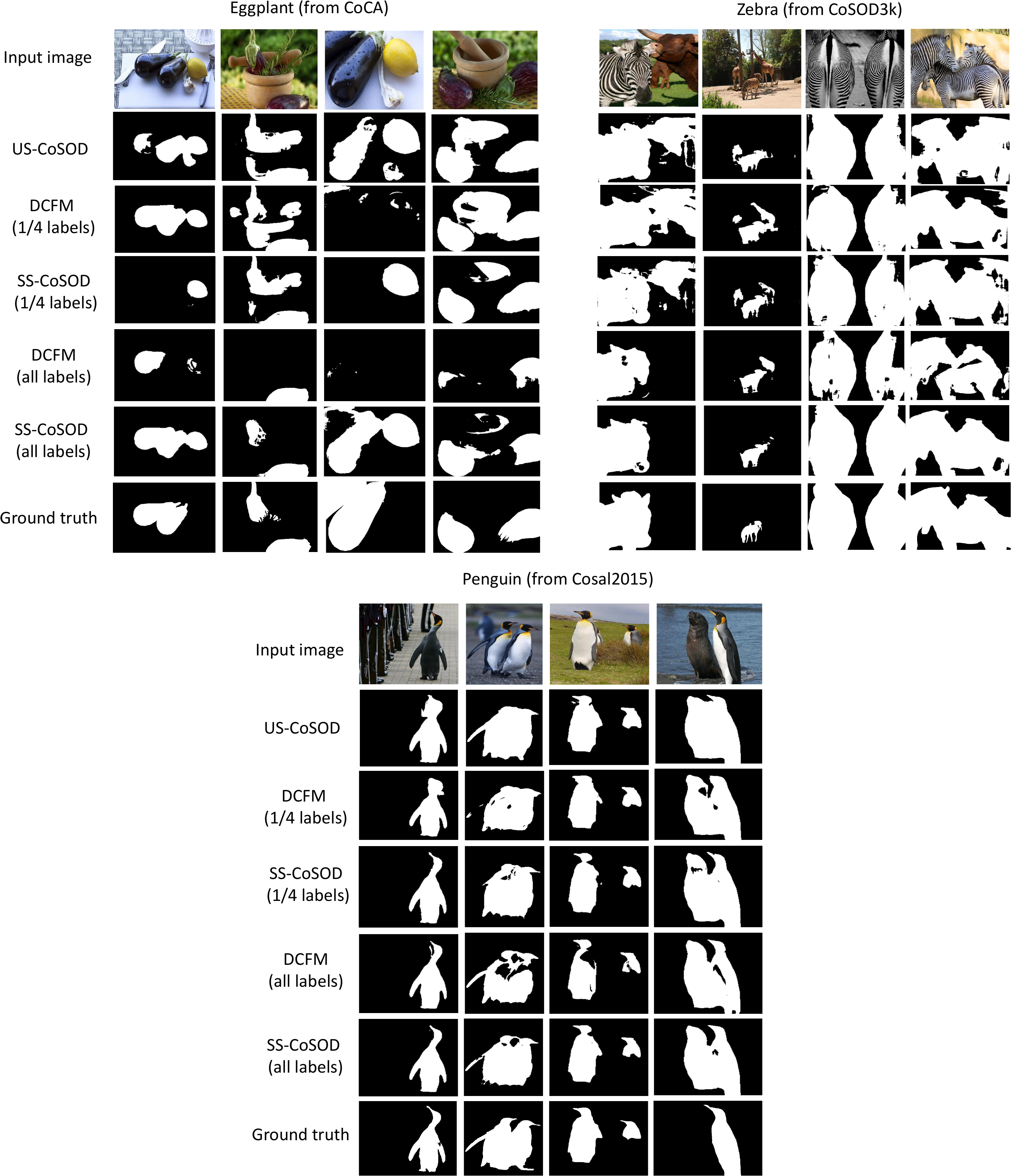}
\caption {Additional qualitative comparisons of our model with different baselines on three image groups selected each from CoCA, CoSOD3k and
Cosal2015. Our SS-CoSOD model trained with all labels produces the most accurate segmentation mask compared to the other baselines.}
\label{fig:quals}
\vspace{-2.3mm}
\end{figure*}

In Fig.~\ref{fig:quals} we present additional results of  co-salient object detection using the proposed models and the other baselines. 

In the first image group in Fig.~\ref{fig:quals} we show the CoSOD predictions on the \textit{eggplant} category from the CoCA dataset. While our US-CoSOD model detects the salient objects well, it fails to accurately segment the eggplant. Similarly, both the DCFM and our SS-CoSOD model  trained with 1/4 labels fail to accurately detect the eggplant instances. Our SS-CoSOD model when trained with all labels predicts CoSOD segmentations most  closely resembling the ground truth. 

In the \textit{zebra} image group (selected from the CoSOD3k dataset), we  observe that the segmentation maps obtained from our SS-CoSOD model trained with all labels most closely resemble the ground truth. The DCFM model trained with all labels suffers from overestimating the co-saliency of certain image regions e.g. in columns 1 and 2 and produces incomplete segmentations in columns 3 and 4. While the DCFM model trained with all labels  segments the zebra in column 1, the segmentation prediction fails to preserve the shape. In column 2, all models except our SS-CoSOD model trained with all labels detect the giraffes as being co-salient along with the zebras. 

\begin{figure*}
\centering
\includegraphics[width = 17.7cm]{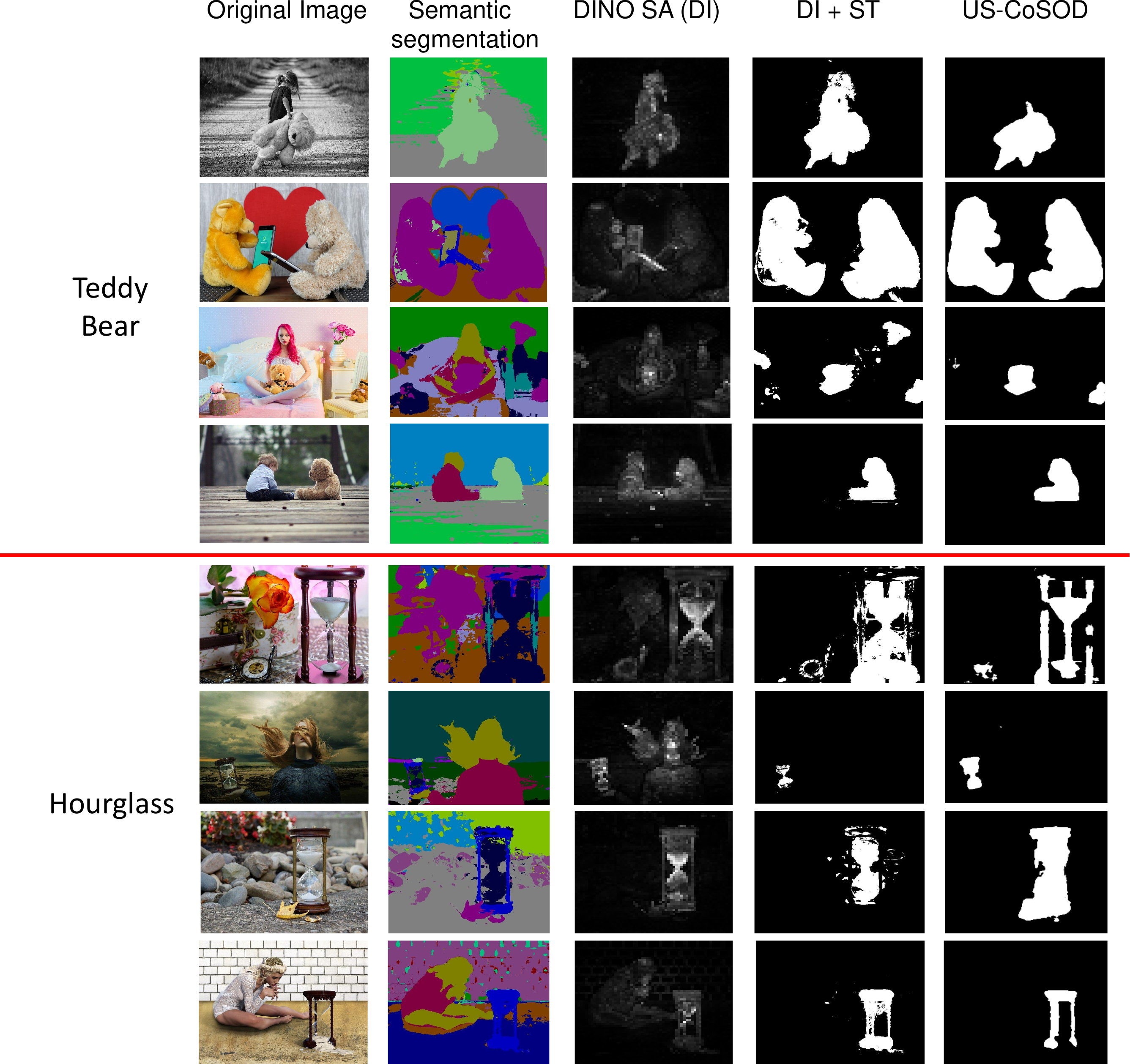}
\caption {Additional qualitative  comparisons of the DINO self-attention  maps (DI), pseudo ground truth co-saliency masks from our DI+ST model, and predictions from our US-CoSOD  model.}
\label{fig:quals2}
\vspace{-2.3mm}
\end{figure*}

In the third image group, we compare the segmentation results on the \textit{penguin} group from the Cosal2015 dataset. Overall, our SS-CoSOD model trained with all labels produces more accurate co-salient object segmentations compared to the other baselines. In column 1, our SS-CoSOD models trained with 1/4 labels and with  all labels well segment the penguin. In the last column of this group, we show an instance where all the models including our SS-CoSOD fail to distinguish the penguin from the seal. This could be due to the fact that the seal has similar visual features as the penguin, which makes it difficult for the models to distinguish between the two categories. Training on more fine-grained categories might help our model resolve this ambiguity.
\subsection{Comparison of unsupervised CoSOD predictions}
In Fig.~\ref{fig:quals2} we present additional results comparing the self-attention (SA) maps from DINO (DI), the pseudo co-salient ground truth masks -  our DINO+STEGO model (DI+ST),  and predictions from our US-CoSOD  model.

In column 2 of row block 1 (the \textit{teddy bear} image group), we observe that the most frequent unsupervised semantic clusters  representing the teddy bear are colored light green and pink. Our US-CoSOD model effectively eliminates the inaccurate segmentation of the child (that carries  the teddy bear) produced by the DINO SA and the DI+ST models. In rows 2 and 3 of this group, US-CoSOD rectifies the inaccurate segmentation masks obtained from the DI+ST model. In row 4, the  teddy bear segmentation from both the DI+ST and US-CoSOD models is quite accurate. 

In row block 2 (the \textit{hourglass} image group), we observe that blue and dark blue colored unsupervised semantic clusters mainly constitute the hourglass object. In row 2 of this group, although the SA map from DINO highlights both the person and the hourglass to be salient, the segmentation predictions from the DI+ST and US-CoSOD models correctly show only the hourglass to be co-salient, which is  due to the fact that the co-occurrence frequency of the unsupervised semantic cluster denoting the hourglass object is sufficiently high compared to that for the person. Our US-CoSOD model further improves the segmentations predicted by  the DI+ST model.

\begin{figure*}
\centering
\includegraphics[width = 14.3cm]{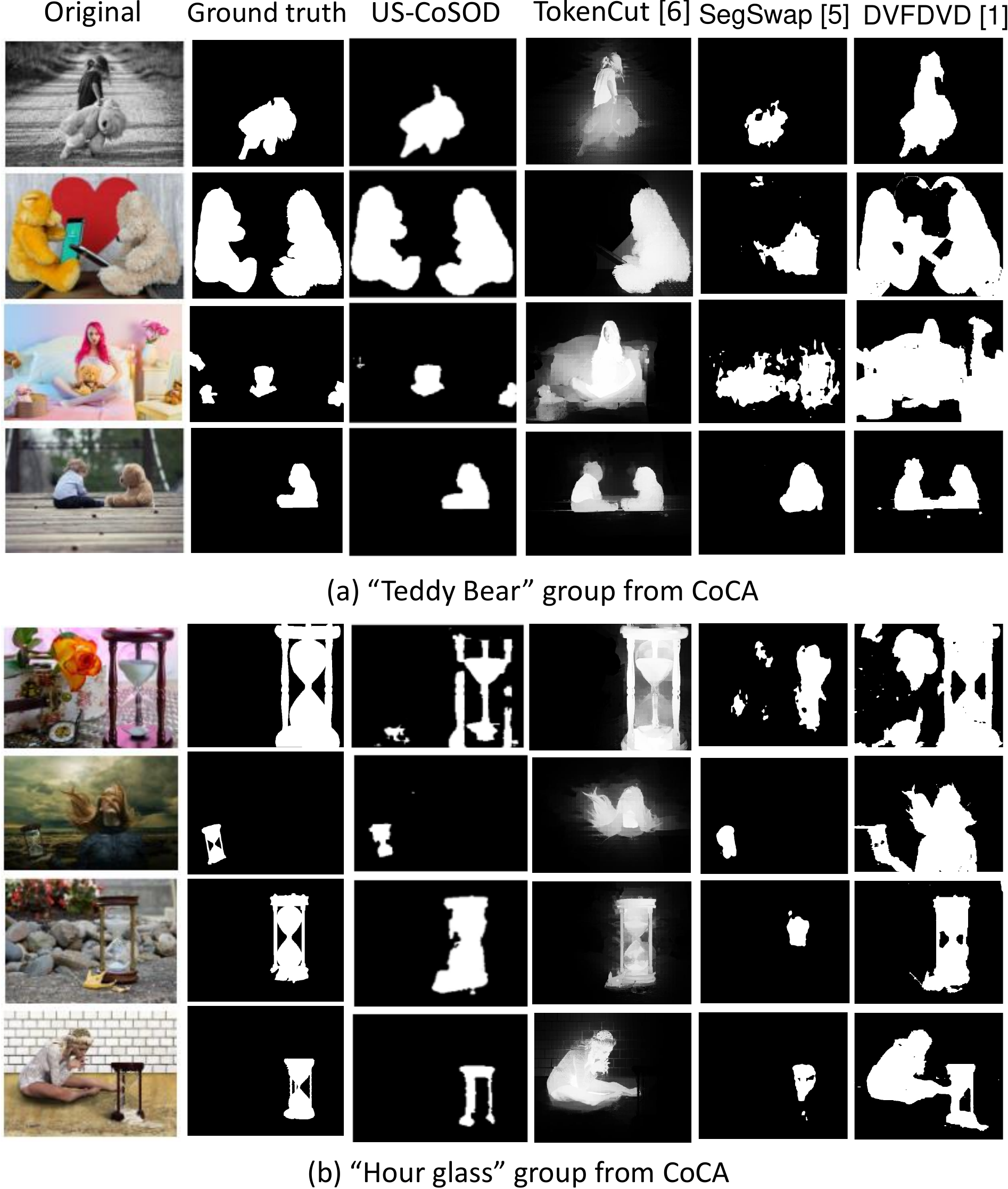}
\caption {Qualitative comparisons of prediction results from our unsupervised CoSOD model, US-CoSOD vs. corresponding segmentations from existing unsupervised segmentation models. TokenCut \cite{wang2022self} is a single-image segmentation method and SegSwap \cite{shen2022learning} and DVFDVD \cite{amir2021deep} are unsupervised co-segmentation methods.}
\label{fig:quals3}
\vspace{-2.3mm}
\end{figure*}

In Fig.~\ref{fig:quals3} we present qualitative  results comparing our method with the different unsupervised methods for single-image segmentation and co-segmentation tasks. We observe that our US-CoSOD model has better segmentation predictions compared to the SegSwap \cite{shen2022learning}, DVFDVD \cite{amir2021deep}, and the TokenCut \cite{wang2022self} models for two image groups - \textit{hour glass} and \textit{teddy bear}  from the CoCA dataset.

\subsection{Confidence Estimation  Network predictions}
\begin{figure*}
\centering
\includegraphics[width = 15.5cm]{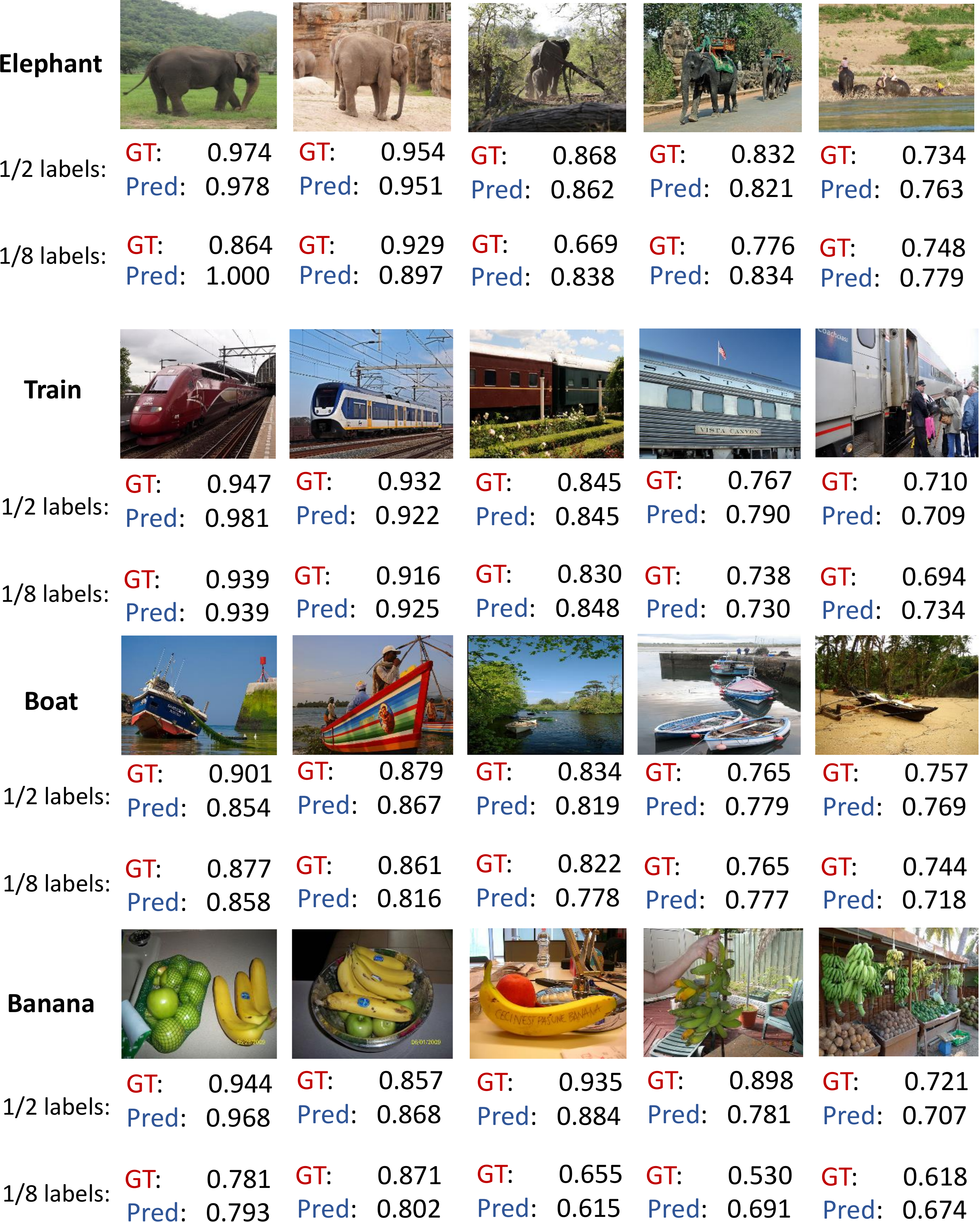}
\caption {Depiction of the ground truth and the predicted confidence scores from our Confidence Estimation Network (CEN) module using 1/2 and 1/8 labels for training. \textit{GT} denotes the max. F1-score of the predictions obtained from the pretrained $f_{PT}$ model (see Fig. 3 in the main paper) on the unlabeled data and \textit{Pred} denotes the F1-score predicted by our trained CEN module. We observe that the confidence scores  vary in proportion to the image complexity in terms of the image contents. Also, we observe that our CEN module is able to predict the ground truth confidence score well.}
\label{fig:confidencer_preds}
\vspace{-2.3mm}
\end{figure*}

In Fig.~\ref{fig:confidencer_preds} we show the ground truth and the predicted confidence scores from our Confidence Estimation Network (CEN) module using 1/2 and 1/8 labels for training. \textit{GT} denotes the max. F1-score of the predictions obtained from the pretrained $f_{PT}$ model (see Fig. 3 in the main paper) on the unlabeled data and \textit{Pred} denotes the F1-score predicted by our trained CEN module. We observe that the confidence scores  vary in proportion to the image complexity in terms of the image contents. In particular, we observe that the ground truth confidence score is high when the co-salient object is more salient and has a clear demarcation with respect to the scene background, while the ground truth confidence score is low when the image scene is more cluttered (e.g. for the \textit{train} class in row 2 and the \textit{banana} class in row 4) or the co-salient objects are out-of-distribution (e.g. for the \textit{boat} class in row 3, the boat is on the land). Also, we observe that our CEN module is able to predict the ground truth confidence score well. Therefore, the CEN model effectively suppresses the error propagation during training caused due to inaccurate confidence estimation on images that are difficult for the CoSOD task.

\section{Additional quantitative results}
\label{sec:intro}

\begin{table*}[t]
\centering
\caption{
Performance evaluation of our US-CoSOD model: we show the prediction performance of US-CoSOD when trained on a set of 50K images (with 50 images per class) along with the other baselines presented in Table 1 in the main paper. US-CoSOD when trained on 150 images per class produces the best performance. 
}
\label{tab:tab1_50class}
\setlength{\tabcolsep}{3pt}
\vspace{1pt}
\resizebox{17.6cm}{!}{
\begin{tabular}{l|llll|llll|llll}
\toprule
 & \multicolumn{4}{c}{CoCA} & \multicolumn{4}{c}{Cosal2015} &
 \multicolumn{4}{c}{CoSOD3k}\\
Method &
MAE$\downarrow$ & $F_{\beta}^{max}\uparrow$ & $E_{\phi}^{max}\uparrow$ & $S_{\alpha}\uparrow$ & MAE$\downarrow$ & $F_{\beta}^{max}\uparrow$ & $E_{\phi}^{max}\uparrow$ & $S_{\alpha}\uparrow$ &
MAE$\downarrow$ & $F_{\beta}^{max}\uparrow$ & $E_{\phi}^{max}\uparrow$ & $S_{\alpha}\uparrow$\\

\midrule
 DINO (DI) &
0.214 & 0.372 & 0.572 & 0.540 & 0.154 & 0.659 & 0.753 & 0.688 & 0.146 & 0.624 & 0.749 & 0.679\\

 STEGO (ST) &
0.235 &  0.353 & 0.555 & 0.523 & 0.164 & 0.618 & 0.717 & 0.676 & 0.204 & 0.543 & 0.660 & 0.615\\

TokenCut \cite{wang2022self} (CVPR 2022) &
0.167 & 0.467 & 0.704 & 0.627 & 
0.139 & 0.805 & 0.857 & 0.793 & 
0.151 & 0.720 & 0.811 & 0.744\\
 
 DVFDVD \cite{amir2021deep} (ECCVW 2022) &
0.223 &  0.422 & 0.592 & 0.581 & 0.092 & 0.777 & 0.842 & 0.809 & 0.104 & 0.722 & 0.819 & 0.773\\

 SegSwap \cite{shen2022learning} (CVPRW 2022) &
0.165 &  0.422 & 0.666 & 0.567 & 0.178 & 0.618 & 0.720 & 0.632 & 0.177 & 0.560 & 0.705 & 0.608\\

Ours (DI+ST) &
0.165 & 0.461 & 0.676 & 0.610 & 
0.112 & 0.760 & 0.823 & 0.767 &
0.124 & 0.684 & 0.793 & 0.724\\

Ours (US-CoSOD-COCO9213)  &
0.140 & 0.498 & 0.702 & 0.641 & 0.090 & 0.792 & 0.852 & 0.806 & 0.095 & 0.735 & 0.832 & 0.772\\

Ours (US-CoSOD-ImgNet50)  &
0.141 & 0.516 & 0.703 & 0.648 &
0.076 & 0.823 & 0.876 & 0.827 &
0.092 & 0.752 & 0.841 & 0.783\\

Ours (US-CoSOD-ImgNet150) &
\textbf{0.116} & \textbf{0.546} & \textbf{0.743} & \textbf{0.672} & \textbf{0.070} & \textbf{0.845} & \textbf{0.886} & \textbf{0.840} & \textbf{0.076} & \textbf{0.779} & \textbf{0.861} & \textbf{0.801}\\

Ours (US-CoSOD-ImgNet450) &
0.127 & 0.543 & 0.726 & 0.666 & 0.071 & 0.844 & 0.884 & 0.842 & 0.079 & 0.775 & 0.854 & 0.800 \\

\bottomrule
\end{tabular}
}
\label{tab:unsup_cosod}
\vspace{-3pt}
\end{table*}

\begin{table*}[t]
\centering
\caption{
Performance comparison of the different versions of our unsupervised and semi-supervised models. In column 1, we indicate the fraction of labeled data for training, followed by the actual number of images. 
}
\label{tab:semisup_cosod}
\setlength{\tabcolsep}{3pt}
\vspace{1pt}
\resizebox{17.8cm}{!}{%
\begin{tabular}{l|l|llll|llll|llll}
\toprule
 & & \multicolumn{4}{c}{CoCA} & \multicolumn{4}{c}{Cosal2015} &
 \multicolumn{4}{c}{CoSOD3k}\\
Split & Method &
MAE$\downarrow$ & $F_{\beta}^{max}\uparrow$ & $E_{\phi}^{max}\uparrow$ & $S_{\alpha}\uparrow$ &
MAE$\downarrow$ & $F_{\beta}^{max}\uparrow$ & $E_{\phi}^{max}\uparrow$ & $S_{\alpha}\uparrow$ &
MAE$\downarrow$ & $F_{\beta}^{max}\uparrow$ & $E_{\phi}^{max}\uparrow$ & $S_{\alpha}\uparrow$ \\

\toprule
  & DCFM \cite{yu2022democracy} (CVPR 22) &
0.119 & 0.485 & 0.725 & 0.636 & 0.088 & 0.780 &  0.847 & 0.786 & 
0.088 & 0.716 &  0.827 &  0.753\\
\cmidrule{2-14}
 &  US-CoSOD+DCFM & 0.108 & 0.557 & 0.754 & 0.683 & 
0.068 & 0.854 & 0.888 & 0.846 & 
0.076 & 0.783 & 0.857 & 0.801\\

  & SS-CoSOD-DJ (w/o CEN) &
0.107 & 0.485 & 0.728 & 0.635
& 0.094 & 0.771 & 0.834 & 0.771 &
0.089 & 0.709 & 0.817 & 0.742\\

1/16 (576)  & SS-CoSOD-DJ (w/ CEN)  &
0.115 & 0.488 & 0.730 & 0.639 &  
0.086 & 0.782 & 0.847 & 0.787 &
0.086 & 0.717 & 0.828 & 0.755 \\

  & SS-CoSOD  &
0.113 & 0.492 & 0.733 & 0.641 &
0.085 & 0.788 & 0.850 & 0.792 & 
0.084 & 0.721 & 0.830 & 0.758 \\

  & US-CoSOD+SS-CoSOD &
0.111 & 0.554 & 0.751 & 0.681 &
\textbf{0.066} & \textbf{0.855} & \textbf{0.890} & \textbf{0.849} &
0.075 & 0.783 & 0.858 & \textbf{0.803}\\

  & SS-CoSOD with ImgNet &
\textbf{0.098} & \textbf{0.562} & \textbf{0.757} & \textbf{0.684} &
0.072 & 0.837 & 0.880 & 0.828 &
\textbf{0.068} & \textbf{0.784} & \textbf{0.865} & 0.800\\

\toprule
 & DCFM \cite{yu2022democracy} (CVPR 22) &
0.110 & 0.493 & 0.731 & 0.639 
& 0.096 & 0.780 & 0.839 &  0.779 &
0.096 & 0.727 & 0.818 & 0.746\\
\cmidrule{2-14}
 &  US-CoSOD+DCFM & 0.111 & 0.558 & 0.754 & 0.683 & 
\textbf{0.067} & 0.857 & \textbf{0.890} & \textbf{0.847} & 
0.076 & 0.785 & 0.857 & 0.801 \\

  & SS-CoSOD-DJ (w/o CEN) &
0.103 & 0.497 & 0.732 & 0.641 &
0.096 & 0.777 & 0.835 & 0.777 &
0.094 & 0.727 & 0.816 & 0.744 \\

1/8 (1152)  & SS-CoSOD-DJ (w/ CEN)  &
0.116 & 0.499 & 0.735 & 0.644  &
0.085 & 0.793 & 0.854 & 0.800 &
0.087 & 0.740 & 0.834 & 0.767 \\

  & SS-CoSOD  &
0.114 & 0.500 & 0.736 & 0.645 &
0.084 & 0.795 & 0.856 & 0.802 &
0.086 & 0.740 & 0.835 & 0.767 \\

  & US-CoSOD+SS-CoSOD &
0.108 & 0.558 & 0.755 & 0.683 &
0.068 & \textbf{0.857} & 0.888 & 0.845 &
0.076 & 0.785 & 0.856 & 0.799  \\

  & SS-CoSOD  with ImgNet  &
\textbf{0.097} & \textbf{0.560} & \textbf{0.755} & \textbf{0.685} & 0.068 & 0.845 & 0.884 & 0.838 & \textbf{0.068} & \textbf{0.791} & \textbf{0.871} & \textbf{0.808}\\

\toprule
 & DCFM \cite{yu2022democracy} (CVPR 22) &
0.107 & 0.547 & 0.758 & 0.672 &
0.073 & 0.829 & 0.880 &  0.824 &
0.075 & 0.775 & 0.862 & 0.794\\
\cmidrule{2-14}
 &  US-CoSOD+DCFM & 0.109 & 0.569 & 0.758 & 0.685 &
0.069 & 0.855 & 0.888 & 0.844 &
0.077 & 0.783 & 0.854 & 0.797\\
 
  & SS-CoSOD-DJ (w/o CEN) &
0.097 & 0.552 & 0.763 & 0.678  &
0.076 & 0.828 & 0.874 & 0.818  &
0.075 & 0.776 & 0.859 & 0.790 \\

 1/4 (2303)  & SS-CoSOD-DJ (w/ CEN) &
0.096 & 0.560 & 0.764 & 0.685  &
0.069 & 0.839 & 0.885 & 0.831 &
0.069 & 0.784 & 0.867 & 0.802  \\

  & SS-CoSOD  &
0.097 & 0.562 & 0.765 & 0.686  &
0.068 & 0.841 & 0.886 & 0.833 &
0.068 & 0.785 & 0.868 & 0.803  \\

  & US-CoSOD+SS-CoSOD &
0.107 & 0.566 & 0.757 & 0.686  &
0.066 & \textbf{0.858} & 0.891 & \textbf{0.848} &
0.073 & 0.787 & 0.859 & 0.803\\

  & SS-CoSOD  with ImgNet  &
\textbf{0.091} & \textbf{0.581} & \textbf{0.772} &  \textbf{0.698} &
\textbf{0.066} & 0.851 & \textbf{0.891} & 0.841  &
\textbf{0.064} & \textbf{0.799} & \textbf{0.875} & \textbf{0.812} \\

\toprule
 &  DCFM \cite{yu2022democracy} (CVPR 22) &
0.101 & 0.566 & 0.764 & 0.690  &
0.065 & 0.845  & 0.889  &  0.838  &
0.070 & 0.792 & 0.870 & 0.807\\
\cmidrule{2-14}
  &  US-CoSOD+DCFM & 0.105 & 0.569 & 0.760 & 0.688 &
0.068 & 0.856 & 0.889 & 0.843 &
0.074 & 0.793 & 0.862 & 0.804\\
 
   & SS-CoSOD-DJ (w/o CEN) &
0.092 & 0.572 & 0.771 & 0.694 &
0.068 & 0.846 & 0.885 & 0.834  &
0.071 & 0.791 & 0.865 & 0.802\\

1/2 (4607)  & SS-CoSOD-DJ (w/ CEN)  &
0.090 & 0.578 & 0.772 & 0.699 &
0.062 & 0.851 & 0.892 & 0.843 &
0.067 & 0.795 & 0.870 & 0.810\\

  & SS-CoSOD  &
0.088 & 0.582 & 0.773 & 0.700  &
0.062 & 0.854 & 0.892 & 0.843  &
0.066 & 0.797 & 0.872 & 0.809 \\

  & US-CoSOD+SS-CoSOD &
0.110 & 0.563 & 0.755 & 0.686 &
0.064 & 0.858 & 0.894 & 0.850 &
0.072 & 0.794 & 0.866 & 0.810 \\

   & SS-CoSOD with ImgNet  &
\textbf{0.088} & \textbf{0.590} & \textbf{0.775} & \textbf{0.705} &
\textbf{0.062} & \textbf{0.861} & \textbf{0.896} & \textbf{0.850} &
\textbf{0.063} & \textbf{0.804} & \textbf{0.876} & \textbf{0.817}\\

\midrule
 & DCFM \cite{yu2022democracy} (CVPR 22) &
\textbf{0.085} & \textbf{0.598} & \textbf{0.783} & \textbf{0.710} &
0.067 & 0.856 & 0.892 & 0.838 &
0.067 & 0.805 & 0.874 & 0.810\\
\cmidrule{2-14}
Full (9213) & US-CoSOD+DCFM & 
0.102 & 0.573 & 0.764 & 0.692 &
0.068 & 0.860 & 0.890 & 0.845 &
0.077 & 0.791 & 0.856 & 0.799\\

 & SS-CoSOD with ImgNet & 0.091 & 0.591 & 0.778 & 0.707 & \textbf{0.061} & \textbf{0.865} & \textbf{0.901} & \textbf{0.852} & \textbf{0.062} & \textbf{0.809} & \textbf{0.882} & \textbf{0.821}\\
 
\midrule
\end{tabular}
}
\vspace{-5pt}
\end{table*}

In Tab.~\ref{tab:tab1_50class}, we provide additional results of the performance evaluation of our US-CoSOD model compared to Tab. 1 in the main paper. In particular, we additionally show the prediction performance of US-CoSOD when trained on a set of 50K images (with 50 images per class) along with the baselines presented in Tab. 1 in the main paper. US-CoSOD when trained on 150 images per class produces the best performance. Training US-CoSOD using 50 images per class (for each of the 1000 ImageNet classes) leads to inferior performance due to limited training data. On the other hand, training the model using 450 images per class reduces segmentation accuracy. This could be because adding more difficult unlabeled images to the training set may lead to erroneous training due to the inaccurate pseudo ground truth masks generated by the DI+ST model, using which US-CoSOD is trained.

\begin{table*}[t]
\centering
\caption{
Comparison of our model with other state-of-the-art models on 3 benchmarks. We achieve state-of-the-art performance on the test datasets.
}
\label{tab:sota_comp}
\setlength{\tabcolsep}{2pt}
\resizebox{17.1cm}{!}{%
\begin{tabular}{l|llll|llll|llll}
\toprule
 & \multicolumn{4}{c}{CoCA} & \multicolumn{4}{c}{Cosal2015} &
 \multicolumn{4}{c}{CoSOD3k}\\

Method &
MAE$\downarrow$ & $F_{\beta}^{max}\uparrow$ & $E_{\phi}^{max}\uparrow$ & $S_{\alpha}\uparrow$ & 
MAE$\downarrow$ & $F_{\beta}^{max}\uparrow$ & $E_{\phi}^{max}\uparrow$ & $S_{\alpha}\uparrow$ &
MAE$\downarrow$ & $F_{\beta}^{max}\uparrow$ & $E_{\phi}^{max}\uparrow$ & $S_{\alpha}\uparrow$\\

\midrule
GCAGC \cite{zhang2020adaptive} (CVPR20)  &
0.111 & 0.523 & 0.754 & 0.669 & 
0.085 & 0.813 & 0.866 & 0.817 &
0.100 & 0.740 & 0.816 & 0.785\\

CoEGNet \cite{fan2021re} (TPAMI21)  &
0.106 & 0.493 & 0.717 & 0.612 &
0.077 & 0.832 & 0.882 & 0.836 &
0.092 & 0.736 & 0.825 & 0.762\\

GICD \cite{zhang2020gradient} (ECCV20) &
0.126 & 0.513 & 0.715 & 0.658 &
0.071 & 0.844 & 0.887 & 0.844 &
0.079 & 0.770 & 0.848 & 0.797\\


GCoNet \cite{fan2021group} (CVPR21) &
0.105 & 0.544 & 0.760 & 0.673 &
0.068 & 0.847 & 0.887 & 0.845 &
0.071 & 0.777 & 0.860 & 0.802 \\

DCFM \cite{yu2022democracy} (CVPR22) &
\textbf{0.085} & 0.598 & \textbf{0.783} & \textbf{0.710} &
0.067 & 0.856 & 0.892 & 0.838 &
0.067 & 0.805 & 0.874 & 0.810 \\

CoRP \cite{zhu2023co} (TPAMI23) &
- & 0.551 & 0.715 & 0.686 &
- & \textbf{0.885} & \textbf{0.913} & \textbf{0.875} &
- & 0.798 & 0.862 & 0.820 \\

UFO \cite{su2023unified} (TMM23) &
0.095 & 0.571 & \textbf{0.782} & \textbf{0.697} &
0.064 & 0.865 & 0.906 & 0.860 &
0.073 & 0.797 & 0.874 & 0.819 \\

GEM \cite{wu2023co} (CVPR23) &
0.095 & 0.599 & \textbf{0.808} & \textbf{0.726} &
0.053 & 0.882 & 0.933 & 0.885 &
0.061 & 0.829 & 0.911 & 0.853 \\

DMT \cite{li2023discriminative} (CVPR23) &
0.108 & \textbf{0.619} & \textbf{0.800} & \textbf{0.725} &
0.0454 & 0.905 & 0.936 & 0.897 &
0.063 & 0.835 & 0.895 & 0.851 \\

Ours (SS-CoSOD with ImgNet) &
0.091 & 0.591 & 0.778 & 0.707 &
\textbf{0.061} & 0.865 & 0.900 & 0.852 &
\textbf{0.062} & \textbf{0.809} & \textbf{0.882} & \textbf{0.821}\\

\bottomrule
\end{tabular}
}
\end{table*}

\begin{table}[t]
\centering
\caption{
Comparison of our US-CoSOD model with a variant version that normalizes the overlap area between the DINO SA mask  and STEGO segmentation mask by the STEGO segmentation mask area. 
}
\label{tab:unsup_ablation}
\setlength{\tabcolsep}{1.5pt}
\resizebox{8.5cm}{!}{%
\begin{tabular}{l|l|llll}
\toprule
Dataset  & Method &
MAE$\downarrow$ & $F_{\beta}^{max}\uparrow$ & $E_{\phi}^{max}\uparrow$ & $S_{\alpha}\uparrow$ \\

\midrule
CoCA & US-CoSOD (Variant) &
 0.131 & 0.410 & 0.650 & 0.590\\
 & US-CoSOD (Proposed) &
 \textbf{0.165} & \textbf{0.461} & \textbf{0.676} & \textbf{0.610} \\
\midrule
Cosal2015 & US-CoSOD (Variant) &
 0.143 & 0.613 & 0.713 & 0.681\\
 & US-CoSOD (Proposed) &
 \textbf{0.112} & \textbf{0.760} & \textbf{0.823} & \textbf{0.767} \\
\midrule
 CoSOD3k & US-CoSOD (Variant) &
 0.127 & 0.579 & 0.714 & 0.666\\
 & US-CoSOD (Proposed) &
 \textbf{0.124} & \textbf{0.684} & \textbf{0.793} & \textbf{0.724} \\

\bottomrule
\end{tabular}
}
\end{table}
\vspace{2mm}
\noindent \textbf{Quantitative evaluation of SS-CoSOD} In Tab.~\ref{tab:semisup_cosod},  we show a more detailed version of Tab. 2 in the main paper. Here, we additionally show the prediction results with 1/8 labeled data.

\vspace{2mm}
\noindent \textbf{Comparison with SOTA} In Tab.~\ref{tab:sota_comp},  we compare the performance of our model with other state-of-the-art models on the 3 benchmark datasets. We outperform the state-of-the-art DCFM model\cite{yu2022democracy} on the Cosal2015 and the CoSOD3k datasets, while we are comparable with this model on the CoCA dataset (DCFM predictions on CoCA being slightly more accurate). Also, we outperform other existing fully supervised CoSOD models by a significant margin.  

\vspace{2mm}
\noindent \textbf{Variant model} In Tab.~\ref{tab:unsup_ablation},  we compare the performance of the proposed US-CoSOD model with a variant version of the model where we divide the overlap area, $O^j_i$ between the DINO mask $DM_i$ and the STEGO segmentation mask $SM^j_i$ (for class $c^j$) by the area occupied by the STEGO mask $Ar(SM^j_i)$. The proposed version of the US-CoSOD model performs better than the variant version over all three test datasets using all four evaluation metrics. 

\vspace{2mm}
\noindent \textbf{Performance on challenging categories} In Tab.~\ref{tab:tab1}, we report the average F-measure score on the categories over which DINO (a pre-trained component) scored lesser than the average DINO F-measure score over the test dataset. Specifically, for a given test dataset, categories that had an F-measure score lower than the threshold value, $F^{\beta}_{th} = \frac{1}{n}\sum_{i=1}^nF^{\beta}(SA_i)$ (here $SA_i$ denotes the DINO self-attention map of image $I_i$ and $n =$ total number of test images) were considered for this experiment. As observed, our US-CoSOD outperforms the pre-trained DINO and DINO+STEGO models by a significant margin on such difficult categories. 

\vspace{2mm}
\noindent \textbf{CEN backbone} In  Tab.~\ref{tab:cen_ablation}, we compared the confidence estimation error (Mean Squared Error) of different backbone networks for our CEN module on the unlabeled dataset. As we observe, ResNet50 trained using DINO provides us the least Mean Squared Error loss across all data splits. We attribute the lower accuracy of MobileNetV2 to its lower feature representation power, and that of the ViTB and ViTS models to the fact that such transformer models fail to outperform convolutional models (\eg~ResNet50) when less data is available for training (in the different label splits).


 


\begin{table}[!h]
\scriptsize
\centering
\caption{Average F-measure of the baselines over the categories on which the categorical F-measure scores of DINO are lower than its average F-measure on the test dataset.}
\label{tab:unsup_compare}
\setlength{\tabcolsep}{2pt}
\resizebox{8.2cm}{!}{%
\begin{tabular}{l|l|l|l}
\toprule
Model & CoCA & Cosal2015 & CoSOD3k \\
\midrule
DINO (DI)  & 0.269 & 0.598 & 0.452 \\
DINO+STEGO (DI+ST)   & 0.331 & 0.654 & 0.529 \\
US-CoSOD  & \textbf{0.408} & \textbf{0.738} & \textbf{0.577} \\
\bottomrule
\end{tabular}}
\vspace{-1mm}
\label{tab:tab1}
\end{table}

\begin{table}[!h]
\scriptsize
\centering
\caption{
Comparison of the confidence estimation error (Mean Squared Error) of different backbone networks for our CEN module on the unlabeled dataset.
}
\label{tab:unsup_compare}
\setlength{\tabcolsep}{2pt}
\resizebox{8.2cm}{!}{%
\begin{tabular}{l|l|l|l}
\toprule
Model & 1/16 (576) & 1/4 (2303) & 1/2 (4607) \\
\midrule
MobileNetV2 (3.4M)  & 0.210 & 0.171 &
 0.168   \\

DINO (ViTS8) (22.2M)   & 0.207 & 0.177 &
 0.174   \\
 
DINO (ViTB8) (86M)  & 0.208 & 0.176 &
0.170  \\

DINO (ResNet50) (25.6M)  & \textbf{0.204} & \textbf{0.166} &
\textbf{0.160}  \\

\bottomrule
\end{tabular}}
\vspace{-1mm}
\label{tab:cen_ablation}
\end{table}

\section{Additional implementation details}
We randomly split the data in the COCO9213 dataset into the labeled and the unlabeled sets (i.e. 1/16, 1/8, 1/4, 1/2 labels) for training the fully supervised DCFM and our semi-supervised SS-CoSOD models. 

The inputs are resized to $224 \times 224$ for both training and inference. We use Adam 
as our optimizer to train our models. The total training time is around 5 hours for US-CoSOD  and around 8 hours for SS-CoSOD using ImageNet-1K. All experiments are run on a single NVIDIA Tesla V100 SXM2 GPU. 

For the unsupervised model (US-CoSOD) and the supervised pre-training on labeled data in stage 1 in Fig. 3 in the main paper, we set the
learning rate is set as $10^{−5}$
for feature extractor and
$10^{−4}$
for other parts, and the weight decay is set as $10^{−4}$, following \cite{yu2022democracy}. Training these models take around 200 epochs using 1/2 and full labels, and around 100 epochs  using 1/4, 1/8, and 1/16 labels. 

For our semi-supervised approach (SS-CoSOD), we fine-tune the pre-trained model (from stage 1 in Fig. 3 in the main paper) using the
learning rate is set as $10^{−7}$
for feature extractor and
$10^{−6}$
for other parts, and the weight decay is set as $10^{−6}$. 

For training our Confidence Estimation  Network, we randomly divided the labeled data into training (80\%) and validation (20\%) sets. We used the Adam optimizer for training with initial learning rate = $2\times10^{−4}$ with a weight decay = $10^{−4}$. The step of the learning rate scheduler is set as 7. We used a batch size of 32 to train this model.


{\small
\bibliographystyle{ieee_fullname}
\bibliography{egbib}
}

\end{document}